\documentclass[11pt]{article}
\usepackage{amsfonts,amsthm,mathrsfs,amssymb,amsmath}
\usepackage{amsmath}
\usepackage[page,title,titletoc,header]{appendix}
\usepackage{color}
\usepackage{authblk}
\usepackage{graphicx}
\usepackage{caption,subcaption}
\usepackage{url}
\usepackage{enumerate,anysize,color}
\usepackage{longtable}

\usepackage[a4paper,hmargin=1in,vmargin=1in]{geometry}

\usepackage{bm}
\newtheorem{thm}{Theorem}[section]
\newtheorem{lemma}[thm]{Lemma}

\newtheorem{corollary}[thm]{Corollary}
\newtheorem{definition}[thm]{Definition}
\newtheorem{rem}[thm]{Remark}
\newtheorem{Exa}{Example}[section]
\newtheorem{as}{Assumption}
\newtheorem{alg}{Algorithm}

\newcommand{\be}{\begin{equation}}
\newcommand{\ee}{\end{equation}}
\newcommand{\bea}{\begin{eqnarray*}}
\newcommand{\eea}{\end{eqnarray*}}
\newcommand{\bflalign}{\begin{flalign*}}
\newcommand{\eflalign}{\end{flalign*}}

\newcommand{\mR}{\mathbb{R}}
\newcommand{\mN}{\mathbb{N}}
\newcommand{\mE}{\mathbb{E}}
\newcommand{\mcE}{\mathcal{E}}

\newcommand{\mcN}{\mathcal{N}}

\newcommand{\bz}{{\bf z}} 
\newcommand{\bx}{{\bf x}} 
\newcommand{\by}{{\bf y}}

\newcommand{\tr}{\operatorname{tr}}
\newcommand{\la}{\langle}
\newcommand{\ra}{\rangle}
\newcommand{\eref}[1] {(\ref{#1})}

\newcommand{\TK}{\mathcal{T}} 

\newcommand{\TKL}{\mathcal{T}_{\PRegPar}} %
\newcommand{\TXL}{\mathcal{T}_{{\bf x}\PRegPar}}

\newcommand{\TKLL}{\mathcal{T}_{\lambda}}

\newcommand{\LK}{\mathcal{L}}

\newcommand{\IK}{\mathcal{S}_{\rho}}
\newcommand{\TX}{\mathcal{T}_{\bf x}}
\newcommand{\SX}{\mathcal{S}_{\bf x}}

\newcommand{\HK}{H}
\newcommand{\HR}{H_{\rho}}
\newcommand{\LR}{L^2_{\rho_X}}

\newcommand{\GL}{\mathcal{G}_{\lambda}} 
\newcommand{\RL}{\mathcal{R}_{\lambda}} 

\newcommand{\FR}{f_{\rho}}
\newcommand{\FH}{f_{\HK}}

\newcommand{\DZF}{\Delta_1} 
\newcommand{\DZS}{\Delta_2} %
\newcommand{\DZT}{\Delta_3} 

\newcommand{\PRegPar}{{\lambda}}

\newcommand{\ESRA}{\omega_{\lambda}^{{\bf z}}} 
\newcommand{\EPSRA}{\omega_{\lambda}} 
\newcommand{\EESRA}{f_{\lambda}^{\bz}}

\newcommand{\Samples}{{\bf z}}

\newcommand{\Outputs}{{\bf y}}

\newcommand{\mcEE}{\widetilde{\mathcal{E}}}

\providecommand{\keywords}[1]{\textbf{\textit{Keywords}} #1}

\begin{document}
  \title{Optimal Rates for Spectral Algorithms with Least-Squares Regression over Hilbert Spaces}

%
  \author{Junhong Lin$^{1}$,  Alessandro Rudi$^{2,3}$,  Lorenzo Rosasco$^{4,5}$, Volkan Cevher$^{1}$  \\
	{ \small $^{1}$Laboratory for Information and Inference Systems, \'{E}cole Polytechnique F\'{e}d\'{e}rale de Lausanne,  CH1015-Lausanne, Switzerland.}\\	
	{ \small $^{2}$	INRIA, Sierra project-team, 75012 Paris, France.}\\
	{ \small $^{3}$	D\'{e}partement Informatique - \'{E}cole Normale Sup\'{e}rieure, Paris, France.} \\
	{ \small $^{4}$ University of Genova, 16146 Genova, Italy.}\\
	{ \small $^{5}$ LCSL,  Massachusetts Institute of Technology and Italian Institute of Technology.}

}

\date{November 5, 2018}
  \maketitle
  \baselineskip 16pt
\begin{abstract}
In this paper, we study regression problems over a separable Hilbert space with the square loss, covering non-parametric regression over a reproducing kernel Hilbert space. We investigate a class of spectral/regularized algorithms, including ridge regression, principal component regression, and gradient methods.
We prove	optimal, high-probability convergence results in terms of variants of norms  for the studied algorithms, considering a capacity assumption on the hypothesis space and a general source condition on the target function. Consequently, we obtain almost sure convergence results with optimal rates. Our results improve and generalize previous results, filling a theoretical gap for the non-attainable cases. 
	
\end{abstract}

\section{Introduction}\label{sec:introduction}

Let the input space $\HK$ be a separable Hilbert space with inner product denoted by
$\la \cdot, \cdot \ra_{\HK}$ and the output space $\mR$. Let $\rho$ be an unknown probability measure on $\HK\times \mR$, $\rho_X(\cdot)$ the induced marginal measure on $\HK$, and  $\rho(\cdot | x)$ the conditional probability measure on $\mR$ with respect to $x \in \HK$ and $\rho$. Let
 the hypothesis space $\HR = \{f: \HK \to  \mR| \exists \omega \in \HK \mbox{ with } f(x) = \la \omega, x \ra_{\HK}, \rho_X \mbox{-almost surely}\}.$ The goal of least-squares regression is to approximately solve the following expected risk minimization,
\be\label{expectedRisk}
\inf_{f \in \HR} \mcE(f), \quad \mcE(f)= \int_{\HK \times \mR} ( f(x) - y)^2 d\rho(x,y),
\ee
where the measure  $\rho$ is  known only through
a sample $\bz =\{z_i=(x_i, y_i)\}_{i=1}^n$ of size $n\in\mN$, independently and identically distributed  according to $\rho$. Let $\LR$ be the Hilbert space of square integral functions from $\HK$ to $\mR$ with respect to $\rho_X$, with its norm given by $\|f\|_{\rho} = \left(\int_{\HK} |f(x)|^2 d \rho_X\right)^{1/2}$.  The function that minimizes the expected risk over all measurable functions is the
regression function \cite{cucker2007learning,steinwart2008support}, defined as
\be\label{regressionfunc}
f_{\rho}(x) = \int_{\mR} y d \rho(y | x),\qquad x \in \HK,  \rho_X\mbox{-almost every}.
\ee
Throughout this paper, we assume that the support of $\rho_X$ is compact and
 there exists a constant $\kappa \in [1,\infty[$, such that
\be\label{boundedKernel} \la x,x' \ra_{\HK} \leq \kappa^2, \quad \forall x,x'\in \HK,   \rho_X\mbox{-almost every}.
\ee
Under this assumption,  $\HR$ is a subspace of $\LR,$ and 
a solution $\FH$ for \eqref{expectedRisk} is the projection of the regression function $f_{\rho}(x)$
onto the closure of $\HR$ in $\LR.$ See e.g., \cite{lin2017optimal,bauer2007regularization}, or Section \ref{sec:learning} for further details.

The above problem was raised for non-parametric regression with kernel methods \cite{cucker2007learning,steinwart2008support} and it is closely related to functional regression \cite{ramsay2006functional}. 
A common and classic approach for the above problem is based on spectral algorithms. It amounts to solving an empirical linear equation, where to avoid over-fitting and to ensure good performance, a filter function for regularization is involved, see e.g., \cite{bauer2007regularization,gerfo2008spectral}. Such approaches include ridge regression, principal component regression, gradient methods and iterated ridge regression.

A large amount of research has been carried out for spectral algorithms within the setting of learning with kernel methods, see e.g., \cite{smale2007learning,caponnetto2007optimal} for Tikhonov regularization, \cite{zhang2005boosting,yao2007early} for gradient methods, and \cite{caponnetto2006,bauer2007regularization} for general spectral algorithms. Statistical results have been developed in these references, but still, they are not satisfactory. For example, most of the previous results either restrict to the case that the space $\HK_{\rho}$ is universal consistency  (i.e.,  $\HR$ is dense in $\LR$) \cite{smale2007learning,yao2007early,caponnetto2006} or the attainable case (i.e., $\FH \in \HR$) \cite{caponnetto2007optimal,bauer2007regularization}. Also, some of these results require an unnatural assumption that the sample size is large enough and the derived convergence rates tend to be (capacity-dependently) suboptimal in the non-attainable cases. Finally, it is still unclear whether one can derive capacity-dependently optimal convergence rates for spectral algorithms under a general source assumption.

In this paper, we study statistical results for spectral algorithms.
Considering a capacity assumption of the space $\HK$ \cite{zhang2006learning,caponnetto2007optimal}, and a general source condition \cite{bauer2007regularization} of the target function $\FH$,  we show high-probability, optimal convergence results in terms of variants of norms for spectral algorithms. As a corollary, we obtain almost sure convergence results with optimal rates.
 The general source condition is used to characterize
the regularity/smoothness of the target function $\FH$ in $\LR$, rather than in $\HR$ as those in \cite{caponnetto2007optimal,bauer2007regularization}. 
The derived convergence rates are optimal in a minimax sense. Our results, not only resolve the issues mentioned in the last paragraph but also generalize previous results to convergence results with different norms and consider a more general source condition. 


\section{Learning with Kernel Methods and Notations} \label{sec:learning}

In this section, we first introduce supervised learning with kernel methods, which is a special instance of the learning setting considered in this paper. We then introduce some useful notations and auxiliary operators.

{\it Learning with Kernel Methods.} 
	Let $\Xi$ be a closed subset of Euclidean space $\mR^d$. Let $\mu$ be an unknown but fixed Borel probability measure on $\Xi \times Y$. Assume that $\mathbf \{(\xi_i, y_i)\}_{i=1}^n$ are i.i.d. from the distribution  $\mu$. A reproducing kernel $K$ is a symmetric function $K: \Xi
	\times \Xi \to \mR$ such that $(K(u_i, u_j))_{i, j=1}^\ell$ is
	positive semidefinite for any finite set of points
	$\{u_i\}_{i=1}^\ell$ in $\Xi$. The kernel $K$ defines a reproducing
	kernel Hilbert space (RKHS) $(\mathcal{H}_K, \|\cdot\|_K)$ as the
	completion of the linear span of the set $\{K_{\xi}(\cdot):=K(\xi,\cdot):
	\xi\in \Xi\}$ with respect to the inner product $\la K_{\xi},
	K_u\ra_{K}:=K(\xi,u).$ For any $f \in \mathcal{H}_K$, the reproducing property holds: $f(\xi) = \la K_{\xi}, f\ra_K.$
	In learning with kernel methods, one considers the following minimization problem
	$$ \inf_{f\in \mathcal{H}_K} \int_{\Xi \times \mR} (f(\xi) - y)^2 d\mu(\xi,y).$$
	Since $f(\xi) = \la K_{\xi},f\ra_{K} $ by the reproducing property, the above can be rewritten as
	$$ \inf_{f\in \mathcal{H}_K} \int_{\Xi \times \mR} (\la f, K_{\xi} \ra_{K} - y)^2 d\mu(\xi,y).$$
	Defining another probability measure
	$\rho(K_{\xi},y) = \mu(\xi,y)$, the above reduces to \eqref{expectedRisk}.

{ \it Notations and Auxiliary Operators.} 
We next introduce some notations and auxiliary operators which will be useful in the following.
For a given bounded operator $L: \LR \to \HK, $ $\|L\|$ denotes the operator norm of $L$, i.e., $\|L\| = \sup_{f\in \LR, \|f\|_{\rho}=1} \|Lf\|_{\HK}$. 

 Let $\IK: \HK \to \LR$ be the linear map $\omega \to \la \omega, \cdot \ra_{\HK}$, which is bounded by $\kappa$ under Assumption \eref{boundedKernel}. Furthermore, we consider the adjoint operator $\IK^*: \LR \to \HK$, the covariance operator $\TK: \HK \to \HK$ given by $\TK = \IK^* \IK$, and the operator $\LK : \LR \to \LR$ given by $\IK \IK^*.$ It can be easily proved that $ \IK^*g = \int_{\HK} x g(x) d\rho_X(x), $ $\LK f = \int_{\HK} f(x) \la x, \cdot \ra_{\HK} d \rho_X(x)$
and $\TK = \int_{\HK} \la \cdot , x \ra_{\HK} x d \rho_X(x).$ 
Under Assumption \eqref{boundedKernel},
the operators $\TK$ and $\LK$ can be proved to be positive trace class operators (and hence compact):
\be\label{eq:TKBound}
\|\LK\| = \|\TK\| \leq \tr(\TK) = \int_{\HK} \tr(x \otimes x)d\rho_X(x) =  \int_{\HK} \|x\|_{\HK}^2 d\rho_{X}(x) \leq \kappa^2.
\ee
For any $\omega \in \HK$,
it is easy to prove the following isometry property \cite{steinwart2008support}
\be\label{isometry}
\|\IK \omega \|_{\rho} = \|\sqrt{\TK} \omega\|_{\HK}.
\ee
Moreover, according to the spectral theorem,
\be\label{eq:rho2hk}
\|\LK^{-{1\over2 }}\IK \omega\|_{\rho} \leq \|\omega\|_{\HK}
\ee

We define the sampling operator $\SX: \HK \to \mR^n$ by $(\SX \omega)_i = \la \omega, x_i \ra_{\HK},$ $i \in [n]$, where the norm $\|\cdot\|_{\mR^n}$ in $\mR^n$ is the Euclidean norm times $1/\sqrt{n}$.
Its adjoint operator $\SX^*: \mR^n \to \HK,$ defined by $\la \SX^*{\bf y}, \omega \ra_{\HK} = \la {\bf y}, \SX \omega\ra_{\mR^n}$ for ${\bf y} \in \mR^n$ is thus given by $\SX^*{\bf y} = {1 \over n} \sum_{i=1}^n y_i x_i.$ Moreover, we can define the empirical covariance operator $\TX: \HK \to \HK$ such that $\TX = \SX^* \SX$. Obviously,
\bea
\TX = {1 \over n} \sum_{i=1}^n \la \cdot, x_i \ra_{\HK} x_i.
\eea
By Assumption \eqref{boundedKernel}, similar to \eref{eq:TKBound}, we have
\be\label{eq:TXbound}
\|\TX\| \leq \tr(\TX) \leq \kappa^2.
\ee

A simple calculation shows that \cite{cucker2007learning,steinwart2008support} for all  $f \in \LR,$
$$
\mcE(f) - \mcE(\FR) = \|f - \FR\|_{\rho}^2.
$$
Then it is easy to see that \eqref{expectedRisk} is equivalent to 
$
\inf_{f\in \HR} \|f - \FR\|_{\rho}^2.
$
Using the projection theorem, one can prove that
a  solution $\FH$ for the above problem is the projection of the regression function $f_{\rho}$ onto the closure of $\HR$ in $\LR$, and moreover,  for all $f\in \HR$, (see e.g., \cite{lin2017optimal}),
\be\label{frFH}
\IK^* f_{\rho} = \IK^* \FH,
\ee
and
\be\label{eq:exceRisk}
\mcE(f) -  \mcE(\FH) = \|f - \FH\|_{\rho}^2.
\ee
\section{Spectral/Regularized Algorithms} \label{sec:spec}

In this section, we demonstrate and introduce spectral algorithms. 

The search for an approximate solution in $\HR$ for Problem \eref{expectedRisk} is equivalent to the search of an approximated solution in $\HK$ for
\be\label{eq:mcEE}
\inf_{\omega \in \HK} \mcEE(\omega),\quad \mcEE(\omega) = \int_{\HK \times \mR} (\la\omega,x \ra_{\HK} - y)^2 d \rho(x,y).
\ee
As the expected risk $\mcEE(\omega)$ can not be computed exactly  and that it can be only approximated through the empirical risk $\mcEE_{\bz}(\omega),$ defined as 
$$\mcEE_{\bz}(\omega) = {1\over n} \sum_{i=1}^n ( \la \omega, x_i \ra_{\HK} - y_i)^2,
$$  a first idea to deal with the problem is to replace the objective function in \eqref{eq:mcEE} with the empirical risk, which leads to an estimator $\hat{\omega}$ satisfying the empirical, linear equation
$$
\TX \hat{\omega} = \SX^* \by.
$$
However, solving the empirical, linear equation directly may lead to a solution that fits the sample points very well but has a large expected risk. This is called as overfitting phenomenon in statistical learning theory. Moreover, the inverse of the empirical covariance operator $\TX$ does not exist in general. To tackle with this issue, a common approach in statistical learning theory and inverse problems, is to replace $\TX^{-1}$ with an alternative, regularized one, which leads to spectral algorithms \cite{engl1996regularization, caponnetto2006,bauer2007regularization}.

A spectral algorithm is generated by a specific choice of filter function.
Recall that the definition of filter functions is given as follows.

\begin{definition}[Filter functions]\label{def}
	Let $\Lambda$ be a subset of $\mR_+.$
	A class of functions $\{\GL: [0, \kappa^2] \to [0,\infty[, \lambda \in \Lambda \}$ is said to be filter functions with qualification $\tau $ ($\tau\geq 1$) if there exist some positive constants $E,F_{\tau}<\infty$ such that
	\be
	\label{eq:GLproper1}
	\sup_{\alpha\in [0,1]} \sup_{\lambda\in \Lambda}\sup_{u \in ]0,\kappa^2] } |u^{\alpha}\GL(u)| \lambda^{1-\alpha} \leq E.
	\ee
	and 
	\be\label{eq:GLproper4}
	\sup_{\alpha\in [0, \tau]} \sup_{\lambda \in \Lambda}	\sup_{u\in]0, \kappa^2]} |(1 - \GL(u)u)|u^{\alpha}\lambda^{-\alpha} \leq F_{\tau} .\ee
\end{definition}

Given a filter function $\GL$,  the spectral algorithm is defined as follows.
\begin{alg}
	\label{alg:dSpe}
	Let $\GL$ be a filter function indexed with   $\lambda >0$. The spectral algorithm over the samples $\bz$ is given by\footnote{ Let $L$ be a self-adjoint, compact operator over a separable Hilbert space $\HK$. $\GL(L)$ is an operator on $L$ defined by spectral calculus: suppose that $\{(\sigma_i, \psi_i)\}_i$ is a set of
		normalized eigenpairs of $L$ with the eigenfunctions $\{\psi_i\}_i$ forming an orthonormal basis
		of $\HK$, then $\GL(\TX) = \sum_i \GL(\sigma_i) \psi_i \otimes \psi_i.$}
	\be\label{eq:ESRA}
	\ESRA = \GL(\TX) \SX^* \by,
	\ee
	and 
	\be\label{eq: EESRA}
	\EESRA = \IK \ESRA.
	\ee
\end{alg}

  Different filter functions correspond to different regularization algorithms. The following examples provide several specific choices on filter functions, which leads to different types of regularization methods, see e.g. \cite{gerfo2008spectral,bauer2007regularization,smale2007learning}.


\begin{Exa}[Spectral cut-off]
	Consider the spectral cut-off or truncated singular value decomposition (TSVD) defined by
	$$
	\GL(u) = 
	\begin{cases}
	u^{-1}, & \mbox{if } u \geq \lambda,\\
	0, & \mbox{if } u < \lambda.
	\end{cases}
	$$
	Then
	the qualification $\tau$ could be any positive number and $E = F_{\tau} = 1$. 
\end{Exa}

\begin{Exa}[Gradient methods]
	The choice
	$\GL(u) = \sum_{k=1}^t \eta(1-\eta u)^{t-k}$ 
with $\eta \in ]0,\kappa^2]$	where we identify $\lambda = (\eta t)^{-1},$
	corresponds to gradient methods  or Landweber iteration algorithm. 
	The qualification $\tau$ could be any positive number, 
	$E =1 ,$ and $F_\tau  = (\tau/\mathrm{e})^{\tau}$. 
\end{Exa}

\begin{Exa}[(Iterated) ridge regression] \label{exa:KRRbc}
	Let $l \in \mN.$
	Consider the function $$\GL(u) = \sum_{i=1}^l \lambda^{i-1} (\lambda+u)^{-i} = {1 \over u} \left( 1 - {\lambda^l \over (\lambda+u)^l}\right).$$ It is easy to show that the qualification $\tau=l$, $E = l$ and $F_{\tau}=1.$
		In the case that $l=1,$ the algorithm is ridge regression.
\end{Exa}
The performance of spectral algorithms can be measured in terms of the excess risk, 
$
\mcE(\EESRA)  - \inf_{\HR} \mcE,
$
which is exactly $ \|\EESRA - \FH\|_{\rho}^2$ according to \eqref{eq:exceRisk}.
Assuming that $\FH \in \HR$, which implies that 
there exists some $\omega_*$ such that $\FH  = \IK \omega_*$ (in this case, the solution with minimal $\HK$-norm for $\FH = \IK \omega$ is denoted by $\omega_{\HK}$), it can be measured in terms of $\HK$-norm,
$
\|\ESRA - \omega_{\HK}\|_{\HK},
$ 
which is closely related to $\|\LK^{-{1\over 2 }}\IK(\ESRA - \omega_{\HK})\|_{\HK} = \|\LK^{-{1\over 2 }}(\EESRA - \FH)\|_{\rho} $ according to \eqref{eq:rho2hk}.
In what follows, we will measure the performance of spectral algorithms in terms of a broader class of norms,
$
\|\LK^{-a}(\EESRA - \FH)\|_{\rho},
$
where $a\in [0,{1\over 2}]$ is such that  $\LK^{-a} \FH$ is well defined. Throughout this paper, we assume that ${1/n}\leq \lambda\leq 1.$

\section{Convergence Results}\label{sec:conve}
In this section, we first introduce some basic assumptions and then present convergence results  for spectral algorithms. 

\subsection{Assumptions}
The first assumption relates to a moment condition on the output value $y$.
\begin{as}\label{as:noiseExp}
	There exists positive constants $Q$ and $M$ such that for all $l \geq 2$ with $l \in \mN,$
	\be\label{noiseExp}
	\int_{\mR} |y| ^{l} d\rho(y|x) \leq {1 \over 2} l! M^{l-2} Q^2, 
	\ee
	$\rho_{ X}$-almost surely. 
\end{as}
The above assumption is very standard in statistical learning theory. It is satisfied if 
$y$ is bounded almost surely, or if $y = \la\omega_*,x\ra_{\HK} + \epsilon$, where  $\epsilon$ is a Gaussian random variable with zero mean and it is independent from $x$.
 Obviously, Assumption \ref{as:noiseExp} implies that the regression function $\FR$
 is bounded almost surely, as
\be\label{eq:bounRegFunc}
|\FR(x)| \leq \int_{\mR} |y| d\rho(y|x) \leq \left(\int_{\mR} |y|^2 d\rho(y|x)\right)^{1\over 2} \leq Q. 
\ee

The next assumption relates to the regularity/smoothness of the target function $\FH.$
 As $\FH \in \overline{Range(\IK)}$ and $\LK = \IK \IK^*,$ it is natural to assume a general source condition on $\FH$ as follows.

\begin{as}\label{as:regularity}
	$\FH$ satisfies  \be\label{eq:FHFR}
\int_{\HK} (\FH(x) - \FR(x))^2 x \otimes x d \rho_X(x) \preceq B^2 \TK,
	\ee 
	and the following source condition
	\be\label{eq:socCon}
	\FH = \phi(\LK) g_0, \quad \mbox{with}\quad  \|g_0\|_{\rho} \leq R.
	\ee
	Here, $B,R \geq 0$ and $\phi:[0,\kappa^2] \to \mR^+$ is a non-decreasing index function such that $\phi(0)=0$ and $\phi(\kappa^2)<\infty$.  
	Moreover,  for some $\zeta \in[0,\tau],$ $\phi(u) u^{-\zeta}$ is non-decreasing, and
	the qualification $\tau$ of $\GL$ covers the index function $\phi$.
\end{as}
Recall that the qualification $\tau$ of $\GL$ covers the index function $\phi$ is defined as follows \cite{bauer2007regularization}.
\begin{definition}\label{def:indFunc}
	We say that the qualification $\tau$ covers the index function $\phi$ if there exists a $c>0$ such that  for all $0<\lambda \leq \kappa^2,$
	\be\label{eq:SCQualfi}
	c {\lambda^{\tau} \over \phi(\lambda)} \leq \inf_{\lambda\leq  u\leq \kappa^2} {u^{\tau} \over 
		\phi(u)}.
	\ee
\end{definition}

Condition \eqref{eq:FHFR} is trivially satisfied if $\FH$ is bounded almost surely. Moreover, when making a consistency assumption, i.e., $\inf_{\HR} \mcE = \mcE(\FR)$, as that in \cite{smale2007learning,caponnetto2006,caponnetto2007optimal,steinwart2009optimal}, for kernel-based non-parametric regression, it is satisfied with $B=0.$
Condition \eref{eq:socCon} is a more general source condition that characterizes the ``regularity/smoothness" of the target function. It is trivially satisfied with $\phi(u) = 1$ as $\FH \in \overline{\HR} \subseteq \LR$. 
 In non-parametric regression with kernel methods, one typically considers 
 H\"{o}lders condition (corresponding to $\phi(u) = u^{\alpha},\alpha\geq 0$) \cite{smale2007learning,caponnetto2007optimal,caponnetto2006} .
 \cite{bauer2007regularization,myleiko2017regularized,rastogi2017optimal} considers a general source condition but only with an index function $\phi(u)\sqrt{u}$, 
 where $\phi$ can be decomposed as $\psi\vartheta$ and $\psi: [0,b] \to \mR_+$ is operator monotone with $\psi(0)=0$ and $\psi(b)<\infty$, and $\vartheta: [0,\kappa^2] \to \mR_+$ is Lipschitz continuous with $\vartheta(0)=0$. In the latter case $\inf_{\HR} \mcE$ has a solution $\FH$ in $\HR$ as that \cite{steinwart2008support,rosasco2015learning}
 \be\label{eq:isoEmb}
 \LK^{{1\over 2}} (\LR) \subseteq \HR,
 \ee
In this paper, we will consider a source assumption with respect to a more general index function, $\phi = \psi \vartheta$, where  $\psi: [0,b] \to \mR_+$ is operator monotone with $\psi(0)=0$ and $\psi(b)<\infty$, and $\vartheta: [0,\kappa^2] \to \mR_+$ is Lipschitz continuous.  Without loss of generality, we assume that the Lipschitz constant of $\vartheta$ is $1$, as one can always scale both sides of the source condition \eqref{eq:socCon}. Recall that the function $\psi$ is called operator monotone on $[0,b]$, if for any pair of self-adjoint operators $U,V$ with spectra in $[0,b]$ such that $U \preceq V$, $\phi(U)\preceq \phi(V).$

Finally, the last assumption relates to the capacity of the hypothesis space $\HR$ (induced by $\HK$).
\begin{as}\label{as:eigenvalues}
	For some $\gamma \in ]0,1]$ and $c_{\gamma}>0$, $\TK$ satisfies
	\be\label{eigenvalue_decay}
	\tr(\TK(\TK+\lambda I)^{-1})\leq c_{\gamma} \lambda^{-\gamma}, \quad \mbox{for all } \lambda>0.
	\ee
\end{as}

The left hand-side of of \eref{eigenvalue_decay} is called as the effective
dimension \cite{caponnetto2007optimal}, or the degrees of freedom \cite{zhang2006learning}.
It can be related to covering/entropy number conditions, see \cite{steinwart2008support} for further details.
Assumption \ref{as:eigenvalues} is always true for $\gamma=1$ and $c_{\gamma} =\kappa^2$, since
$\TK$ is a trace class operator which implies the eigenvalues of $\TK$, denoted as $\sigma_i$, satisfy
$\tr(\TK) = \sum_{i} \sigma_i \leq \kappa^2.$
This is referred to as the capacity independent setting.
Assumption \ref{as:eigenvalues} with $\gamma \in]0,1]$ allows to derive better  rates. It is satisfied, e.g.,
if the eigenvalues of $\TK$ satisfy a polynomial decaying condition $\sigma_i \sim i^{-1/\gamma}$, or with $\gamma=0$ if $\TK$ is finite rank.

\subsection{Main Results}
Now we are ready to state our main results as follows.

\begin{thm}\label{thm}
	Under Assumptions \ref{as:noiseExp}, \ref{as:regularity} and \ref{as:eigenvalues}, let $a \in [0,{1\over 2}\wedge \zeta]$, $\PRegPar = n^{\theta-1}$ with $\theta\in[0,1]$, and 
	$\delta\in]0,1[.$ The followings hold with probability at least $1-\delta$.
	\\
	1) If $\phi: [0,b] \to \mR_+$ is operator monotone 
	with  $b> \kappa^2$,  and $\phi(b) <\infty$, or Lipschitz continuous with constant $1$ over $[0,\kappa^2]$, then 
	\begin{align}\label{eq:mainErrBoun1}
	&\|\LK^{-a} ( \EESRA  - \FH) \|_{\rho} \nonumber\\
	\leq&  \lambda^{-a}\left({\tilde{C}_1\over n\lambda^{{1\over 2}\vee (1-\zeta)}}  + {\tilde{C}_2 \over \sqrt{n\lambda^{\gamma}}} + \tilde{C}_3 \phi(\lambda)\right) \log{6\over \delta}\left(\log{6\over \delta}+ \gamma(\theta^{-1} \wedge \log n)\right)^{1-a}.
	\end{align}
	2) If $\phi = \psi \vartheta$, where $\psi:[0,b] \to \mR_+$ is operator monotone
	with  $b> \kappa^2$, $\psi(0) = 0$ and $\psi(b) < \infty$, and $\vartheta:[0,\kappa^2] \to \mR_+$ is non-decreasing, Lipschitz continuous with constant $1$ and $\vartheta(0)=0$. Furthermore, assume that the quality of $\GL$ covers $\vartheta(u)u^{{1\over 2}-a},$
	then
	\begin{align}\label{eq:mainErrBoun2}
	\|\LK^{-a} ( \EESRA  - \FH) \|_{\rho} 
	\leq&  \lambda^{-a}\log{6\over \delta}\left(\log{6\over \delta}+ \gamma(\theta^{-1} \wedge \log n)\right)^{1-a} \\
	& \ \times 
	\left({\tilde{C}_1\over n\lambda^{{1\over 2}\vee (1-\zeta)}}  + {\tilde{C}_4 \over \sqrt{n\lambda^{\gamma}}} + \tilde{C}_5 \phi(\lambda) + \tilde{C}_6 \vartheta(\lambda)\psi(n^{-{1\over 2}})\right), \nonumber
	\end{align}
	Here, $\tilde{C}_1,\tilde{C}_2,\cdots,\tilde{C}_6$ are positive constants depending only on 
	$\kappa^2,c_{\gamma},\gamma,\zeta, \phi,\tau$ $B,M,Q,R,E,F_{\tau},$ $b,a,c$ and $\|\TK\|$  (independent from $\PRegPar,n,\delta,$ and $\theta$, and given explicitly in the proof).
\end{thm}

The above theorem provides convergence results with respect to variants of norms in high-probability  for spectral algorithms.
 Balancing the different terms in the upper bounds, one has the following results with an optimal, data-dependent choice of regularization parameters. Throughout the rest of this paper,  $C$ is denoted as a positive constant that depends only on $\kappa^2,c_{\gamma},\gamma,\zeta, \phi,\tau$ $B,M,Q,R,E,F_{\tau},$ $b,a,c$ and $\|\TK\|$, and it could be different at its each appearance.
\begin{corollary}\label{cor:generalSoucr}
	Under the assumptions and notations of Theorem \ref{thm}, let $2\zeta+\gamma >1$ and $\lambda = \Theta^{-1}(n^{-1})$ where $\Theta(u) = (\phi(u)/\phi(1))^2u^{\gamma}.$
	The following holds with probability at least $1-\delta.$ \\
	1) Let $\phi$ be as in Part 1) of Theorem \ref{thm}, then
		\begin{align}\label{eq:errBouGEn}
	\|\LK^{-a} ( \EESRA  - \FH) \|_{\rho} 
	\leq  C {\phi(\Theta^{-1}(n^{-1})) \over (\Theta^{-1}(n^{-1}))^a} \log^{2-a}{6\over \delta}.
	\end{align}
	2) Let $\phi$ be as	in Part 2) of Theorem \ref{thm} and $\lambda \geq  n^{-{1\over 2}}$, then \eqref{eq:errBouGEn} holds.
	\end{corollary}
The error bounds in the above corollary are optimal as they match the minimax rates from \cite{rastogi2017optimal} (considering only the case $\zeta \geq 1/2$ and $a=0$). The assumption that the quality of $\GL$ covers $\vartheta(u) u^{1\over 2}$ in Part 2) of Corollary \ref{cor:generalSoucr}  is also implicitly required in   \cite{bauer2007regularization,myleiko2017regularized,rastogi2017optimal}, and it is always satisfied for principle component analysis and gradient methods. The condition $\lambda\geq n^{-1/2}$ will be satisfied  in most cases when  the index function has a Lipschitz continuous part, and moreover, it is trivially satisfied when $\zeta \geq 1,$ as will be seen from the proof.

As a direct corollary of Theorem \ref{thm}, we have the following results considering H\"{o}lder source conditions.
\begin{corollary}\label{cor}
	Under the assumptions and notations of Theorem \ref{thm}, we let $\phi(u) = \kappa^{-2(\zeta-1)_+} u^{\zeta} $ in Assumption \ref{as:regularity} and $\lambda = n^{-{1\over 1\vee(2\zeta+\gamma)}}$, then with probability at least $1-\delta,$
	\be\label{eq:hold}
		\|\LK^{-a} (\EESRA  - \FH) \|_{\rho} \leq C\begin{cases} n^{-{\zeta-a \over 2\zeta+\gamma}} \log^{2-a}{6\over \delta} & \mbox{ if } 2\zeta+\gamma>1, \\
			n^{-(\zeta-a)} \log{6\over \delta} \left(\log{6\over \delta} + \log n^{\gamma}\right)^{1-a}
			& \mbox{ if } 2\zeta+\gamma\leq 1.
		\end{cases}
	\ee
\end{corollary}
The error bounds in \eqref{eq:hold} are optimal as the convergence rates match the minimax rates shown in \cite{caponnetto2007optimal,blanchard2016optimal}  with $\zeta \geq 1/2$. The above result asserts that spectral algorithms with an appropriate regularization parameter converge optimally.

Corollary \ref{cor} provides convergence results in high-probability for the studied algorithms.
It implies convergence in expectation and almost sure convergence shown in the follows. Moreover, when $\zeta \geq 1/2,$ it can be translated into convergence results with respect to norms related to $\HK.$

\begin{corollary}\label{cor:3}
Under the assumptions of Corollary \ref{cor}, the following holds.\\
	1) For any $q \in \mN_+,$ we have  
		\be
	\mE\|\LK^{-a} (\EESRA  - \FH) \|_{\rho}^q \leq C\begin{cases} n^{-{q(\zeta-a) \over 2\zeta+\gamma}}  & \mbox{ if } 2\zeta+\gamma>1, \\
		n^{-q(\zeta-a)}  \left(1 \vee \log n^{\gamma}\right)^{q(1-a)}
		& \mbox{ if } 2\zeta+\gamma\leq 1.
	\end{cases}
	\ee
	2) For any $0<\epsilon<{\zeta-a}$,
	$$
	\lim_{n\to\infty} {\|\LK^{-a} (\IK \EESRA  - \FH) \|_{\rho}  n^{\zeta-a -\epsilon \over 1\vee(2\zeta+\gamma)}} = 0, \quad \mbox{almost surely}.
	$$
	3) If $\zeta \geq 1/2,$ then for some $\omega_{\HK} \in \HK,$ $\IK \omega_{\HK} = \FH$ almost surely, and with probability at least $1-\delta,$
		\be
	\|\TK^{{1\over 2} - a} ( \ESRA  - \omega_{\HK}) \|_{\HK} \leq C n^{-{\zeta-a \over 2\zeta+\gamma}} \log^{2-a}{6\over \delta}.
	\ee
	\end{corollary}

\begin{rem}
	If $\HK = \mR^d,$ then Assumption \ref{as:eigenvalues} is trivially satisfied with $c_{\gamma} = \kappa^2 (d \wedge \sigma_{\min}^{-1}),\gamma=0$,
	and Assumption \ref{as:regularity} could be satisfied \footnote{Note that this is not true in general if $\HK$ is a general Hilbert space, and the proof for the finite-dimensional cases could be simplified, leading to some smaller constants in the error bounds.} with any $\zeta>1/2$. Here $\sigma_{\min}$ denotes the smallest eigenvalue of $\TK$. 
	Thus, following from the proof of Theorem \ref{thm}, we have that with probability at least $1-\delta,$
	$$
	\|\LK^{-a} (\EESRA  - \FH) \|_{\rho}  \leq C \sqrt{c_{\gamma} \over n} \log{6\over \delta} \left(\log{6\over \delta} \log c_{\gamma} \right)^{1-a}.
	$$
\end{rem}
The proof for all the results stated in this subsection are postponed in the next section.
\subsection{Discussions} 
There is a large amount of research on  theoretical results  for non-parametric regression with kernel methods in the literature, see e.g.,
\cite{wu2006learning,rudi2015less,szabo2015two,lin2017distributed,dicker2017kernel,myleiko2017regularized,rudi2017generalization,lin2018optimal}  and references therein.  As noted in Section \ref{sec:learning}, our results apply to non-parametric regression with kernel methods.
In what follows, we will translate some of the results for kernel-based regression into results for regression over a general Hilbert space and compare our results with  these results.

We first compare Corollary \ref{cor} with some of these results in the literature for spectral algorithms with H\"{o}lder source conditions. Making a source assumption as
\be\label{eq:HSCB}
 \FR = \LK^{\zeta} g_0 \quad \mbox{with } \|g_0\|_{\rho} \leq R, 
\ee
$1/2 \leq \zeta \leq \tau$, and with $\gamma>0$,
 \cite{guo2017learning} shows that with probability at least $1-\delta,$
$$
	\|\EESRA  - \FR \|_{\rho} \leq C n^{-{\zeta\over 2\zeta+\gamma}} \log^{4}{1\over \delta}.
$$
Condition \eqref{eq:HSCB} implies that $\FR \in \overline{\HR}$ as $\HR = range(\IK)$ and $\LK = \IK \IK^*$. Thus $\FH = \FR $ almost surely.\footnote{Such a assumption is satisfied  if  $\inf_{\HR}\mcEE = \mcEE(\FR)$ and it is supported by many function classes and reproducing kernel Hilbert space in learning with kernel methods \cite{steinwart2008support}.}
In comparison, Corollary \ref{cor} is more general. It provides convergence results in terms of different norms for  a more general H\"{o}lder source condition, allowing $0<\zeta \leq 1/2$ and $\gamma=0.$ Besides, it does not require the extra assumption $\FH = \FR$ and the derived error bound in \eqref{eq:hold} has a  smaller depending order on $\log {1\over \delta}.$
For the assumption \eqref{eq:HSCB} with $0\leq \zeta<1/2$, certain results are derived in 
\cite{smale2007learning} for Tikhonov regularization and  in \cite{yao2007early} for gradient methods, but the rates are suboptimal and capacity-independent. Recently, \cite{lin2018optimal} shows that under the assumption \eqref{eq:HSCB}, with $\zeta\in]0,\tau]$ and $\gamma\in [0,1]$, spectral algorithm has the following error bounds in expectation, 
	\bea
\mE\| \EESRA  - \FR\|_{\rho}^2 \leq C \begin{cases} n^{-{2\zeta \over 2\zeta+\gamma}} & \mbox{ if } 2\zeta+\gamma>1, \\
	n^{-2\zeta} (1 \vee \log n^{\gamma})
	& \mbox{ if } 2\zeta+\gamma\leq 1.
\end{cases}
\eea
Note also that \cite{dicker2017kernel} provides the same optimal error bounds as the above, but only restricts to the cases $ 1/2\leq \zeta\leq \tau$ and $n \geq n_0$.
In comparison,  Corollary \ref{cor} is more general. It  provides convergence results with different norms and it does not require the universal consistency assumption. The derived error bound in \eqref{eq:hold} is more meaningful as it holds with high probability.  However,
it has an extra logarithmic factor in the upper bound for the case $2\zeta+\gamma\leq1,$ which is worser than that from \cite{lin2018optimal}.
\cite{bauer2007regularization,blanchard2016optimal} study statistical results  for spectral algorithms, under a H\"{o}lder source condition, 
$\FH \in \LK^{\zeta} g_0$ with $1/2\leq \zeta \leq \tau.$ Particularly,  \cite{blanchard2016optimal} shows that if \be
\label{eq:sampLarge}n \geq C {\lambda^{-2}}\log^2{1\over \delta},
\ee
then with probability at least $1-\delta$, with $1/2<\zeta\leq \tau$ and $0\leq a\leq 1/2$,
	\bea
\|\LK^{-a} (\EESRA  - \FH) \|_{\rho} \leq C n^{-{\zeta-a \over 2\zeta+\gamma}} \log{6\over \delta}.
\eea
In comparison, Corollary \ref{cor} provides optimal convergence rates even in the case that $0\leq \zeta\leq 1/2$, while it does not require the extra condition \eqref{eq:sampLarge}.
Note that we do not pursue an error bound that depends both on $R$ and the noise level as those in \cite{blanchard2016optimal,dicker2017kernel}, but it should be easy to modify our proof to derive such error bounds (at least in the case that $\zeta \geq 1/2$). The only results by now for the non-attainable cases  with a general H\"{o}lder condition with respect to $\FH$ (rather than $\FR$) are from \cite{lin2017optimal}, where convergence rates of order $O(n^{-{\zeta \over 1\vee (2\zeta+\gamma)}} \log^2 n)$ are derived (but only) for gradient methods assuming $n$ is large enough.

We next compare Theorem \ref{thm} with results from \cite{bauer2007regularization,rastogi2017optimal} for spectral algorithms considering general source conditions.  Assuming that $\FH \in \phi(\LK) \sqrt{\LK} g_0$ with $\|g_{0}\|_{\rho} \leq R$ (which implies $\FH = \IK \omega_{\HK}$ for some $\omega_{\HK} \in \HK$,) where $\phi$ is as in Part 2) of Theorem \ref{thm},
\cite{bauer2007regularization} shows that if the qualification of $\GL$ covers $\phi(u)\sqrt{u}$ and 
\eqref{eq:sampLarge} holds,
then with probability at least $1-\delta,$ 
$$
\|\LK^{-a} (\EESRA - \FH)\|_{\rho} \leq C \lambda^{-a} \left(  \phi(\lambda)\sqrt{\lambda}  + {1\over \sqrt{\lambda n}} \right) \log{6 \over \delta}, \quad a=0,{1\over 2}.
$$
The error bound is capacity independent, i.e., with $\gamma=1$. Involving the capacity assumption\footnote{Note that from the proof from \cite{rastogi2017optimal}, we can see the results from \cite{rastogi2017optimal} also require \eqref{eq:sampLarge}.}, the error bound is further improved in \cite{rastogi2017optimal}, to 
$$
\|\LK^{-a} (\EESRA - \FH)\|_{\rho} \leq C \lambda^{-a} \left(  \phi(\lambda)\sqrt{\lambda}  + {1\over \sqrt{n\lambda^{\gamma} }} \right) \log{6 \over \delta},\quad a=0,{1\over 2}.
$$
As noted in \cite[Discussion]{guo2017learning}, these results lead to the following estimates in expectation 
$$
\mE\|\LK^{-a} (\EESRA - \FH)\|_{\rho}^2 \leq C \lambda^{-2a} \left(  \phi(\lambda)\sqrt{\lambda}  + {1\over \sqrt{n\lambda^{\gamma} }} \right)^2 \log n, \quad a=0,{1\over 2}
$$
In comparison with these results, Theorem \ref{thm} is more general, considering a general source assumption and covering the general case that $\FH$ may not be  in $\HR$. Furthermore, it provides convergence results with respect to a broader class of norms, and
 it does not require the condition \eqref{eq:sampLarge}. 
 Finally, it leads to convergence results in expectation with a better rate (without the logarithmic factor) when the index function is $\phi(u)\sqrt{u}$, and it can infer almost-sure convergence results.

%

\section{Proofs} \label{sec:proof}
In this section, we prove the results stated in  Section \ref{sec:conve}.  We first give some basic lemmas, and then give the proof of the main results.

\subsection{Lemmas}

{\it Deterministic Estimates}\\
We first introduce the following lemma, which is a generalization of \cite[Proposition 7]{bauer2007regularization}. For notational simplicity, we denote \be\label{eq:residual}\RL(u) = 1 - \GL(u)u,\ee
and 
$$
\mcN(\lambda) = \tr(\TK (\TK + \lambda)^{-1}).
$$
\begin{lemma}\label{lem:sorConReslt}
	Let $\phi:[0,\kappa^2] \to \mR^+$ be a non-decreasing index function and the qualification $\tau$ of the filter function $\GL$ covers the index function $\phi$, and for some $\zeta \in[0,\tau],$ $\phi(u) u^{-\zeta}$ is non-decreasing. Then for all $a \in[0,\zeta],$
	\be	\sup_{0< u \leq \kappa^2} | \RL(u)| \phi(u) u^{-a} \leq c_g \phi(\lambda)\lambda^{-a}, \quad c_g = {F_{\tau} \over c \wedge 1}, \label{eq:sorConReslt}
	\ee where $c$ is from Definition \ref{def:indFunc}.
\end{lemma}
\begin{proof}
	When $ \lambda \leq u \leq \kappa^2$, by \eref{eq:SCQualfi},  we have
	$$
	{\phi(u) \over u^{\tau}} \leq {1\over c} {\phi(\lambda) \over \lambda^{\tau}}.
	$$
	Thus,
	$$
	|\RL(u) | \phi(u) u^{ -a} =  |\RL(u) | u^{\tau-a} \phi(u) u^{ -\tau} \leq 	|\RL(u)| u^{\tau-a}  c^{-1} \phi(\lambda) \lambda^{-\tau} \leq F_{\tau} c^{-1} \lambda^{-a} \phi(\lambda),
	$$
	where for the last inequality, we used \eref{eq:GLproper4}.
	When $0<u \leq \lambda,$ since $\phi(u)u^{-\zeta}$ is non-decreasing,
	$$|\RL(u)| \phi(u) u^{-a} = |\RL(u) |u^{\zeta-a}  \phi(u) u^{- \zeta} \leq 
	| \RL(u)|u^{\zeta-a} \phi(\lambda) \lambda^{-\zeta} \leq
	F_{\tau} \phi(\lambda) \lambda^{-a},$$
	where we used \eref{eq:GLproper4} for the last inequality.
	From the above analysis, one can finish the proof.
	\end{proof}

Using the above lemma, we have the following results for the deterministic vector $\EPSRA$, defined by 
\be\label{eq:popFunc}
\EPSRA = \GL(\TK) \IK^* \FH.
\ee

\begin{lemma}
	Under Assumption \ref{as:regularity}, we have for all $ a \in[0, \zeta],$
	\be\label{eq:trueBias}
	\|\LK^{-a}(\IK \EPSRA - \FH)\|_{\rho} \leq c_gR \phi(\lambda) \lambda^{-a} ,
	\ee
	and 
	\be\label{eq:popSeqNorm}
	\|\EPSRA\|_{\HK} \leq E \phi(\kappa^2) \kappa^{-(2\zeta \wedge 1)} \lambda^{-({{1\over2} - \zeta})_+}.
	\ee
\end{lemma}
The left hand-side of 
\eqref{eq:trueBias} is often called as the true bias.
\begin{proof}
	Following from the definition of $\EPSRA$ in \eref{eq:popFunc}, we have
	$$
	\IK \EPSRA - \FH = \IK \GL(\TK) \IK^* \FH - \FH = (\LK\GL(\LK) - I)\FH.
	$$
	Introducing with \eref{eq:socCon}, with the notation $\RL(u) = 1 - \GL(u)u,$ we get
	$$
	\|\LK^{-a}(\IK \EPSRA - \FH)\|_{\rho} = \|\LK^{-a} \RL(\LK) \phi(\LK) g_0 \|_{\rho} \leq \|\LK^{-a} \RL(\LK) \phi(\LK)\| R.
	$$
	Applying the spectral theorem with \eqref{eq:TKBound} and Lemma \ref{lem:sorConReslt} which leads to
	$$
	\|\LK^{-a} \RL(\LK) \phi(\LK)\| \leq \sup_{u\in[0,\kappa^2]} |\RL(u)| u^{-a}\phi(u) \leq c_g \phi(\lambda) \lambda^{-a},
	$$
	one can get \eqref{eq:trueBias}. \\
	From the definition of $\EPSRA$ in \eqref{eq:popFunc} and applying \eqref{eq:socCon}, we have
	$$
\|\EPSRA\|_{\HK} = \|\GL(\TK) \IK^* \phi(\LK) g_0\|_{\HK} \leq \|\GL(\TK) \IK^* \phi(\LK)\| R.
	$$
	According to the spectral theorem, with \eqref{eq:TKBound}, one has 
	$$
	\|\GL(\TK) \IK^* \phi(\LK)\| =\sqrt{	\| \phi(\LK) \IK \GL(\TK)\GL(\TK) \IK^* \phi(\LK)\|} = \|\GL(\LK) \LK^{1\over 2}\phi(\LK)\| \leq \sup_{u \in [0,\kappa^2]} |\GL(u) u^{1\over 2} \phi(u)|.
	$$
	Since both $\phi(u)$ and $\phi(u)u^{-\zeta}$  are non-decreasing and non-negative over $[0,\kappa^2]$,  thus $\phi(u) u^{-\zeta'}$ is also non-decreasing for any $\zeta' \in [0, \zeta].$ 
	If $\zeta\geq 1/2,$ then
	$$
\sup_{u \in [0,\kappa^2]} 	|\GL(u)| u^{1\over 2} \phi(u) = \sup_{u \in [0,\kappa^2]} 	|\GL(u)| u \phi(u) u^{-{1\over 2}}  \leq E\phi(\kappa^2) \kappa^{-1},$$
where for the last inequality, we used \eqref{eq:GLproper1} and that $\phi(u)u^{-{1\over 2}}$ is non-decreasing.
If $\zeta< 1/2,$ similarly, we have
	$$
	\sup_{u \in [0,\kappa^2]} 	|\GL(u)| u^{1\over 2} \phi(u) = 
\sup_{u \in [0,\kappa^2]} 	|\GL(u)| u^{{1\over 2}+\zeta} \phi(u) u^{-\zeta} \leq  	
	E \lambda^{\zeta-{1\over 2}} \phi(\kappa^2) \kappa^{-2\zeta}.$$
	From the above analysis, one can prove \eqref{eq:popSeqNorm}.
	\end{proof}

\noindent{\it Probabilistic Estimates} \\
We next introduce the following lemma, whose prove can be found in \cite{lin2018optimal}.
Note that the lemma improves those from \cite{hsu2014random} for the matrix cases and
Lemma 7.2 in \cite{rudi2013sample} for the operator cases , as it does not need the assumption that the sample size is large enough while considering the influence of $\gamma$ for the logarithmic factor.

\begin{lemma}\label{lem:operDifRes}
	Under Assumption \ref{as:eigenvalues},
	let $\delta\in(0,1)$, $\lambda= n^{-\theta}$ for some $\theta\geq 0$, and
\be\label{eq:aa}
	a_{n,\delta,\gamma}(\theta) =  8\kappa^2 \left(\log{ {4\kappa^2(c_{\gamma}+1) }\over \delta \|\TK\|} + \theta \gamma \min\left({1 \over \mathrm{e}(1-\theta)_+},\log n\right)\right).
	\ee
	We have with probability at least $1-{\delta},$
	\bea
	\| (\TK+\lambda)^{1/2}(\TX+\lambda)^{-1/2}\|^2 \leq 3 a_{n,\delta,\gamma}(\theta) (1 \vee n^{\theta-1}),
	\eea
	and 
	\bea
	\| (\TK+\lambda)^{-1/2}(\TX+\lambda)^{1/2}\|^2 \leq {4\over 3} a_{n,\delta,\gamma}(\theta) (1 \vee n^{\theta-1}),
	\eea
\end{lemma}

To proceed the proof of our next lemmas,
we need the following concentration result for Hilbert space valued random variable
used in \cite{caponnetto2007optimal} and based on the results in \cite{pinelis1986remarks}.

\begin{lemma}
	\label{lem:Bernstein}
	Let $w_1,\cdots,w_m$ be i.i.d random variables in a Hilbert space with norm $\|\cdot\|$. Suppose that
	there are two positive constants $B$ and $\sigma^2$ such that
	\be\label{bernsteinCondition}
	\mE [\|w_1 - \mE[w_1]\|^l] \leq {1 \over 2} l! B^{l-2} \sigma^2, \quad \forall l \geq 2.
	\ee
	Then for any $0< \delta <1/2$, the following holds with probability at least $1-\delta$,
	$$ \left\| {1 \over m} \sum_{k=1}^m w_m - \mE[w_1] \right\| \leq 2\left( {B \over m} + {\sigma \over \sqrt{ m }} \right) \log {2 \over \delta} .$$
\end{lemma}

The following lemma is a consequence of the lemma above (see e.g., \cite{smale2007learning} for a proof).

\begin{lemma}\label{lem:statEstiOper}
	Let $0<\delta<1/2.$ It holds with probability at least $1-\delta:$
	\bea
	\|\TK-\TX\| \leq \|\TK - \TX\|_{HS} \leq { 6 \kappa^2 \over \sqrt{{n}}} \log {2\over \delta}.
	\eea
	Here, $\|\cdot\|_{HS}$ denotes the Hilbert-Schmidt norm.
\end{lemma}

 One novelty of this paper is the following new lemma, which provides a probabilistic estimate on the terms caused by both the variance and approximation error. The lemma is mainly motivated by \cite{smale2007learning,caponnetto2007optimal,lin2017optimal,lin2018optimal}.
 Note that the condition \eqref{eq:FHFR} is slightly weaker than the condition $\|\FH\|_{\infty}<\infty$ required  in \cite{lin2017optimal} for analyzing gradient methods.

\begin{lemma}\label{lem:samAppErr}
	Under Assumptions \ref{as:noiseExp}, \ref{as:regularity} and \ref{as:eigenvalues}, let  $\EPSRA$ be given by \eref{eq:popFunc}. For all $\delta \in ]0,1/2[,$ the following holds with probability at least $1-{\delta}:$
	\be
	\|\TKLL^{-1/2} [(\TX \EPSRA - \SX^* \Outputs) - (\TK \EPSRA - \IK^* \FR)] \|_{\HK} 
	\leq   \left({C_1 \over n \lambda^{{1\over 2} \vee (1-\zeta)}} + \sqrt{ {C_2 (\phi(\lambda))^2\over n\lambda} + {C_3 \over n \lambda^{\gamma}} }\right) 
\log{2\over \delta}.
	\ee
	Here, $C_1 = 8\kappa(M + E\phi(\kappa^2) \kappa^{(1-2\zeta)_+}), C_2 = 96 c_g^2 R^2 \kappa^2$ and $C_3 = 32 (3B^2+ 4Q^2)c_{\gamma}.$
\end{lemma}

\begin{proof}
	Let $\xi_i = \TKLL^{-{1\over 2}} (\la \EPSRA, x\ra_{\HK} - y_i) x_i$ for all $i \in [n].$ 
	From the definition of the regression function $\FR$ in \eref{regressionfunc} and \eqref{frFH}, a simple calculation shows that
	\be\label{eq:interm1}
	\mE[\xi] = \mE[\TKLL^{-{1\over 2}} (\la \EPSRA, x\ra_{\HK} - \FR(x)) x] = \TKLL^{-{1\over 2}}(\TK \EPSRA - \IK^* \FR) =  \TKLL^{-{1\over 2}}(\TK \EPSRA - \IK^* \FH).
	\ee
	In order to apply Lemma \ref{lem:Bernstein}, we need to estimate $\mE[\|\xi - \mE[\xi]\|_{\HK}^l]$ for any $l \in \mN$ with $l\geq 2.$
	In fact, using H\"{o}lder's inequality twice,
	\begin{align}\label{eq:interm6}
	\mE\|\xi - \mE[\xi]\|_{\HK}^l  \leq  \mE\left(\|\xi\|_{\HK}+ \mE\|\xi\|_{\HK}\right)^l \leq 2^{l-1} (\mE \|\xi\|_{\HK}^l + (\mE\|[\xi]\|_{\HK})^l) \leq 2^{l} \mE \|\xi\|_{\HK}^l. 
		\end{align}
	We now estimate $\mE \|\xi\|_{\HK}^l.$ By H\"{o}lder's inequality,
	\begin{align*}
	\mE \|\xi\|_{\HK}^l = \mE [\|\TKLL^{-{1\over 2}}x\|_{\HK}^l (y- \la\EPSRA,x\ra_{\HK})^l]  \leq 2^{l-1}\mE [\|\TKLL^{-{1\over 2}}x\|_{\HK}^l (|y|^l +  |\la \EPSRA,x\ra_{\HK}|^l)].
		\end{align*}
		According to \eqref{boundedKernel}, one has 
		\be\label{eq:interm2}
		\|\TKLL^{-{1\over 2}} x\|_{\HK} \leq \|\TKLL^{-{1\over 2}}\| \|x\|_{\HK} \leq {1\over \sqrt{\lambda}} \kappa.\ee Moreover, by Cauchy-Schwarz inequality and \eqref{boundedKernel}, $|\la \EPSRA, x \ra_{\HK} |\leq \|\EPSRA\|_{\HK} \|x\|_{\HK} \leq \kappa \|\EPSRA\|_{\HK} .$ Thus, we get
			\begin{align}\label{eq:interm5}
		\mE \|\xi\|_{\HK}^l  \leq 2^{l-1} \left({\kappa \over \sqrt{\lambda}} \right)^{l-2}\mE [\|\TKLL^{-{1\over 2}}x\|_{\HK}^2 (|y|^l + (\kappa \|\EPSRA\|_{\HK})^{l-2} |\la\EPSRA,x\ra_{\HK}|^2).
		\end{align}
		Note that by \eqref{noiseExp},
		\begin{align*}
		\mE [\|\TKLL^{-{1\over 2}}x\|_{\HK}^2 |y|^l ] =& \int_{\HK} \|\TKLL^{-{1\over 2}}x\|_{\HK}^2 \int_{\mR} |y|^l d\rho(y|x) d\rho_X(x) \\
		\leq & {1\over 2} l! M^{l-2} Q^2 \int_{\HK} \|\TKLL^{-{1\over 2}}x\|_{\HK}^2 d\rho_X(x). 
		\end{align*}
		Using $\|w\|_{\HK}^2 = \tr(w\otimes w)$ which implies 
		\be\label{eq:interm3}
		 \int_{\HK}  \|\TKLL^{-{1\over 2}}x\|_{\HK}^2 d\rho_X(x) = \int_{\HK}  \tr(\TKLL^{-{1\over 2}}x\otimes x \TKLL^{-{1\over 2}}) d\rho_X(x) = \tr(\TKLL^{-{1\over 2}}\TK \TKLL^{-{1\over 2}})  = \mcN(\lambda),
		\ee
		we get
		\begin{align}\label{eq:interm4}
		\mE [\|\TKLL^{-{1\over 2}}x\|_{\HK}^2 |y|^l ] 
		\leq  {1\over 2} l! M^{l-2} Q^2 \mcN(\lambda). 
		\end{align}
	Besides, by Cauchy-Schwarz inequality, $$
	\mE [\|\TKLL^{-{1\over 2}}x\|_{\HK}^2 |\la\EPSRA,x\ra_{\HK}|^2] \leq 3	\mE [\|\TKLL^{-{1\over 2}}x\|_{\HK}^2(|\la\EPSRA,x\ra_{\HK} - \FH(x)|^2 + |\FH(x) - \FR(x)|^2 +|\FR(x)|^2 )].$$ 
	By \eqref{eq:interm2} and \eqref{eq:trueBias},
	$$ 
		\mE [\|\TKLL^{-{1\over 2}}x\|_{\HK}^2(|\la\EPSRA,x\ra_{\HK} - \FH(x)|^2] \leq {\kappa^2 \over \lambda} \mE[|\la\EPSRA,x\ra_{\HK} - \FH(x)|^2] = {\kappa^2 \over \lambda} \|\IK \EPSRA - \FH\|_{\rho}^2 \leq c_{g}^2R^2\kappa^2 {(\phi(\lambda))^2 \over \lambda},
	$$
	and by \eqref{eq:bounRegFunc} and \eqref{eq:interm3}, 
	$$\mE [\|\TKLL^{-{1\over 2}}x\|_{\HK}^2 |\FR(x)|^2] \leq Q^2 \mE [\|\TKLL^{-{1\over 2}}x\|_{\HK}^2] = Q^2 \mcN(\lambda).
	$$
	Therefore, 
	$$
	\mE [\|\TKLL^{-{1\over 2}}x\|_{\HK}^2 |\la\EPSRA,x\ra_{\HK}|^2] \leq 3\left(c_g^2R^2\kappa^2\phi^2(\lambda)\lambda^{-1} + 
		\mE [\|\TKLL^{-{1\over 2}}x\|_{\HK}^2 |\FH(x) - \FR(x)|^2] +Q^2 \mcN(\lambda) \right).$$ 
	Using $\|w\|_{\HK}^2 = \tr(w\otimes w)$  and \eqref{eq:FHFR}, we have
\begin{align*}
\mE [\|\TKLL^{-{1\over 2}}x\|_{\HK}^2 |\FH(x) - \FR(x)|^2]  = & \mE [ |\FH(x) - \FR(x)|^2\tr(\TKLL^{-{1\over 2}}x\otimes x \TKL^{-{1\over 2}})] \\
=&  \tr(\TKLL^{-1}\mE [(\FH(x) - \FR(x))^2 x\otimes x] ) \\
\leq& B^2\tr(\TKLL^{-1} \TK) = B^2\mcN({\lambda}),
\end{align*}
and therefore,
$$
\mE [\|\TKLL^{-{1\over 2}}x\|_{\HK}^2 |\la\EPSRA,x\ra_{\HK}|^2] \leq 3\left(c_g^2R^2\kappa^2(\phi(\lambda))^2\lambda^{-1} + 
(B^2+Q^2) \mcN(\lambda) \right).$$ 
Introducing the above estimate and \eqref{eq:interm4} into \eqref{eq:interm5}, we derive
\begin{align*}
\mE\|\xi\|_{\HK}^l \leq& 2^{l-1} \left( {\kappa \over \sqrt{\lambda}} \right)^{l-2}\left( {1\over 2} l! M^{l-2} Q^2 \mcN(\lambda) + 3(\kappa \|\EPSRA\|_{\HK})^{l-2}(c_g^2 R^2 \kappa^2 (\phi(\lambda))^2\lambda^{-1} + (B^2+ Q^2)\mcN(\lambda)) \right) \\
\leq & 2^{l-1} \left( {\kappa M + \kappa^2\|\EPSRA\|_{\HK}\over \sqrt{\lambda}} \right)^{l-2} {1\over 2} l! \left(  Q^2 \mcN(\lambda) + 3(c_g^2 R^2 \kappa^2 (\phi(\lambda))^2\lambda^{-1} + (B^2+ Q^2)\mcN(\lambda)) \right),\\
\leq & 2^{l-1} \left( {\kappa M + \kappa^2\|\EPSRA\|_{\HK}\over \sqrt{\lambda}} \right)^{l-2} {1\over 2} l! \left(   3c_g^2 R^2 \kappa^2 (\phi(\lambda))^2\lambda^{-1} + (3B^2+ 4Q^2)c_{\gamma} \lambda^{-\gamma} \right),
\end{align*}
where for the last inequality, we used Assumption \ref{as:eigenvalues}.
Introducing the above estimate into \eref{eq:interm6}, and then substituting with \eref{eq:popSeqNorm} and noting that $\lambda\leq 1,$
we get
$$
\mE[\|\xi - \mE[\xi]\|_{\HK}^l] \leq {1\over 2} l! \left( {4\kappa (M + E\phi(\kappa^2) \kappa^{(1-2\zeta)_+}) \over \lambda^{{1\over 2} \vee (1-\zeta)}} \right)^{l-2} 8\left(  3c_g^2 R^2 \kappa^2 (\phi(\lambda))^2\lambda^{-1} + (3B^2+ 4Q^2)c_{\gamma} \lambda^{-\gamma} \right).
$$
Applying Lemma \ref{lem:Bernstein}, one can get the desired result.
	\end{proof}

\noindent{\it Basic Operator Inequalities} 

\begin{lemma} \cite[Cordes inequality]{fujii1993norm}
	\label{lem:operProd}
	Let $A$ and $B$ be two positive bounded linear operators on a separable Hilbert space. Then
	\bea
	\|A^s B^s\| \leq \|AB\|^s, \quad\mbox{when } 0\leq s\leq 1.
	\eea
\end{lemma}

\begin{lemma}\label{lem:operMon} \cite{mathe2002moduli,mathe2006regularization}
	Suppose $\psi$ is an operator monotone index function on $[0,b],$ with $b>1$. Then there is a constant $c_{\psi}<\infty$ depending on $b-a$, such that for any pair $B_1, B_2,$ $\|B_1\|, \|B_2\| \leq a$, of non-negative self-adjoint operators on some Hilbert space, it holds,
	$$
	\|\psi(B_1) - \psi(B_2)\| \leq c_{\psi} \psi(\|B_1 - B_2\|).
	$$
	Moreover, there is $c_{\psi}'>0$ such that
	$$
	c_{\psi}' {\lambda\over  \psi(\lambda)} \leq {\sigma \over \psi(\sigma)},
	$$
	whenever $0< \lambda < \sigma \leq a <b.$
\end{lemma}

\begin{lemma}\label{lem:operLips}
Let $\vartheta: [0,a] \to \mR_+$ be Lipschitz continuous with constant $1$ and $\vartheta(0) = 0.$ Then for any pair $B_1, B_2,$ $\|B_1\|, \|B_2\| \leq a$, of non-negative self-adjoint operators on some Hilbert space, it holds,
$$
\|\vartheta(B_1) - \vartheta(B_2)\|_{HS} \leq \|B_1 - B_2\|_{HS}.
$$
 \end{lemma}
\begin{proof}
	The result follows from \cite[Subsection 8.2]{birman2003double}.
	\end{proof}

\subsection{Proof of Main Results}
Now we are ready to prove Theorem \ref{thm}.
\begin{proof}[Proof of Theorem \ref{thm}]
		Following from Lemmas \ref{lem:operDifRes}, \ref{lem:statEstiOper} and \ref{lem:samAppErr}, and by a simple calculation, with $\PRegPar=n^{\theta-1}$ and $\theta\in[0,1],$
	we get that with probability at least $1-\delta,$ the following holds:
	\be\label{eq:statEstA}
	\|\TKL^{1\over 2} \TXL^{-{1\over 2}}\|^2 \vee \|\TKL^{-{1\over 2}} \TXL^{{1\over 2}}\|^2 \leq \DZF,\quad \|\TK - \TX\| \leq  \|\TK - \TX\|_{HS} \leq \DZT,
	\ee
	\be\label{eq:statEstB}
	\mbox{and}\quad \|\TKLL^{-1/2} [(\TX \EPSRA - \SX^* \Outputs) - (\TK \EPSRA - \IK^* \FR)] \|_{\HK} \leq \DZS ,
	\ee
	where
	$$\DZF = C_4 (\log {6\over \delta} + \gamma({1\over \theta}\wedge \log n)), \quad C_4 = 24 \kappa^2 \log {2 \mathrm{e}\kappa^2(c_{\gamma}+1) \over \|\TK\|},
	$$
	$$\DZS = \left({C_1 \over n \lambda^{{1\over 2} \vee (1-\zeta)}} + 
	\sqrt{C_2} {\phi(\lambda)} + 
	{\sqrt{C_3} \over \sqrt{n \lambda^{\gamma}} }\right) 
	\log{6\over \delta},
	$$
	$$
	\DZT = {6\kappa^2\over \sqrt{n}} \log {6\over \delta}.
	$$
	Obviously, we have $\DZF \geq 1$ since $\log{6 \over \delta} >1$ and
	 by \eqref{eq:TKBound}, $C_4 \geq 1$. 
	
	We now begin with  the following inequality:
	\bea
\|\LK^{-a} (\EESRA  - \FH) \|_{\rho} = \|\LK^{-a} (\IK \ESRA  - \FH) \|_{\rho} \leq \|\LK^{-a}\IK (\ESRA - \EPSRA)\|_{\rho} + \|\LK^{-a}(\IK \EPSRA - \FH)\|_{\rho}. 
	\eea
	Introducing with \eqref{eq:trueBias}, we get
		\begin{align*}
	\|\LK^{-a} (\EESRA  - \FH) \|_{\rho} \leq& \|\LK^{-a}\IK (\ESRA - \EPSRA)\|_{\rho} + c_g R\phi(\lambda) \lambda^{-a} \\
	\leq & \|\LK^{-a}\IK \TK^{ a-{1\over 2}}\|  \|\TKL^{{1\over 2}-a} \TXL^{a-{1\over 2}}\|  \| \TXL^{{1\over 2}-a} (\ESRA - \EPSRA)\|_{\HK}  + c_g R\phi(\lambda) \lambda^{-a}.
	\end{align*}
By the spectral theorem, $\LK = \IK \IK^*$, $\TK = \IK^* \IK,$ and \eref{eq:TKBound},
 we have 
$\|\LK^{-a} \IK \TK^{a- {1\over 2}} \| \leq 1. $
Moreover, by Lemma \ref{lem:operProd} and $0\leq a  \leq  \zeta \wedge {1\over 2},$
$$
\|\TKL^{{1\over 2}-a} \TXL^{a-{1\over 2}}\| = \|\TKL^{{1\over 2}(1-2a)} \TXL^{-{1\over 2} (1-2a)}\| \leq \|\TKL^{{1\over 2}} \TXL^{-{1\over 2} }\|^{1-2a} \leq \DZF^{{1\over 2} - a}.
$$
We thus get
	\begin{align*}
\|\LK^{-a} (\EESRA  - \FH) \|_{\rho} 
\leq  \DZF^{{1\over 2} - a} \| \TXL^{{1\over 2}-a} (\ESRA - \EPSRA)\|_{\HK}  + c_g R\phi(\lambda) \lambda^{-a}.
\end{align*}
Subtracting and adding with the same term, using the triangle inequality and recalling the notation $\RL(u) $ defined in \eqref{eq:residual}, we get
	\begin{align*}
\|\LK^{-a} (\EESRA  - \FH) \|_{\rho} 
\leq  \DZF^{{1\over 2} - a} \left(\| \TXL^{{1\over 2} -a}\RL(\TX)  \EPSRA\|_{\HK} + \|\TXL^{{1\over 2} -a} (\ESRA - \GL(\TX) \TX \EPSRA ) \|_{\HK} \right)  + c_g R\phi(\lambda) \lambda^{-a}.
\end{align*}
Introducing with \eref{eq:ESRA},
\begin{align}\label{eq:binterm}
\|\LK^{-a} (\EESRA  - \FH) \|_{\rho} 
\leq  \DZF^{{1\over 2} - a} \left(\| \TXL^{{1\over 2} -a}\RL(\TX)  \EPSRA\|_{\HK} + \|\TXL^{{1\over 2} -a} \GL(\TX)(\SX^*\by- \TX \EPSRA ) \|_{\HK} \right)  + c_g R\phi(\lambda) \lambda^{-a}.
\end{align}
{\bf Estimating $ \|\TXL^{{1\over 2}-a} \GL(\TX)(\SX^* \Outputs -  \TX \EPSRA ) \|_{\HK}$:} \\
We first have
\begin{align*}
\|\TXL^{{1\over 2}-a} \GL(\TX)(\SX^* \Outputs -  \TX \EPSRA ) \|_{\HK} \leq& \|\TXL^{{1\over 2}-a} \GL(\TX)\TXL^{1\over 2}\| \|\TXL^{-{1\over 2}} \TKL^{1\over 2}\|
\|\TKL^{-{1\over 2}}(\SX^* \Outputs -  \TX \EPSRA ) \|_{\HK} .
\end{align*}
With \eqref{eq:GLproper1} and \eqref{eq:TXbound}, we have
\bea
\|\TXL^{{1\over 2}-a} \GL(\TX)\TXL^{1\over 2}\| \leq \sup_{u \in [0,\kappa^2]} |(u + \PRegPar)^{1-a} \GL(u)| \leq  \sup_{u \in [0,\kappa^2]} |(u^{1-a} + \PRegPar^{1-a}) \GL(u)| \leq 2 E \lambda^{-a} ,
\eea
 and thus
\begin{align*}
\|\TXL^{{1\over 2}-a} \GL(\TX)(\SX^* \Outputs -  \TX \EPSRA ) \|_{\HK}  
\leq&  2E \lambda^{-a} \DZF^{1/2} \|\TKL^{-{1\over 2}}(\SX^* \Outputs -  \TX \EPSRA ) \|_{\HK} 
\end{align*}
Since by \eqref{eq:trueBias} and \eqref{frFH},
\begin{align*}
&\|\TKL^{-{1\over 2}}(\SX^* \Outputs -  \TX \EPSRA ) \|_{\HK} \\
\leq & \|\TKL^{-{1\over 2}}[(\SX^* \Outputs -  \TX \EPSRA)- (\TK \EPSRA- \IK^* \FR) ]\|_{\HK} + \|\TKL^{-{1\over 2}} (\TK \EPSRA - \IK^* \FR)\|_{\HK} \\
\leq &   \|\TKL^{-{1\over 2}}[(\SX^* \Outputs -  \TX \EPSRA)- (\TK \EPSRA - \IK^* \FR) ]\|_{\HK} + \|\TKL^{-{1\over 2}}\IK^*\| \|\IK \EPSRA -  \FH\|_{\rho}\\
\leq & \DZS + c_g R\phi(\lambda),
\end{align*}
we thus have
\begin{align}\label{eq:term1}
\|\TXL^{{1\over 2}-a} \GL(\TX)(\SX^* \Outputs -  \TX \EPSRA ) \|_{\HK} 
\leq  2E  \lambda^{-a} \DZF^{1/2} ( \DZS + c_g R\phi(\lambda)).
\end{align}
{\bf Estimating $\| \TXL^{{1\over 2} -a}\RL(\TX)  \EPSRA\|_{\HK}$: }\\ 
Note that from the definition of $\EPSRA$ in \eqref{eq:popFunc}, \eqref{eq:socCon},
 $\LK = \IK\IK^*$, and $\TK = \IK^* \IK,$
 $$\EPSRA = \GL(\TK) \IK^*\phi(\LK) g_0 = \GL(\TK) \phi(\TK) \IK^* g_0 ,$$
and thus,
 \begin{align}\label{eq:term2Interm}
 \| \TXL^{{1\over 2} - a}\RL(\TX)  \EPSRA\|_{\HK}
 \leq \| \TXL^{{1\over 2} - a}\RL(\TX)\GL(\TK) \phi(\TK) \IK^* \| R = \| \TXL^{{1\over 2} - a}\RL(\TX)\GL(\TK) \phi(\TK) \TK^{1\over 2} \| R .
 \end{align}
In what follows, we will estimate $\| \TXL^{{1\over 2} - a}\RL(\TX)\GL(\TK) \phi(\TK) \TK^{1\over 2} \|$, considering three different cases.\\
{\it Case 1: $\phi(\cdot)$ is  operator monotone.} \\
We first have
\begin{align*}
\| \TXL^{{1\over 2} -a }\RL(\TX) \phi(\TK)  \GL(\TK)  \TK^{1\over 2}\| \leq& \| \TXL^{{1\over 2} -a } \RL(\TX) \TXL^{1\over 2}\| \|\TXL^{-{1\over 2}} \TKL^{1\over 2}\| \|\TKL^{-{1\over 2}}\TK^{1\over 2}\| \|\phi(\TK) \GL(\TK)\| \\
\leq &  \| \TXL^{{1} -a } \RL(\TX) \| \DZF^{1\over 2}  \|\phi(\TK) \GL(\TK)\|
\end{align*}
By the spectral theorem and \eqref{eq:GLproper4}, with \eqref{eq:TXbound}, 
\bea
\| \TXL^{{1} -a } \RL(\TX) \| \leq \sup_{u \in [0,\kappa^2]} |(u+\PRegPar)^{1-a} \RL(u)|
\leq \sup_{u \in [0,\kappa^2]} |(u^{1-a}+\PRegPar^{1-a}) \RL(u)| \leq 2F\lambda^{1-a},
\eea
(where we write $F_{\tau} = F$ throughout)
and it thus follows that
\begin{align*}
\| \TXL^{{1\over 2} -a }\RL(\TX) \phi(\TK)  \GL(\TK)  \TK^{1\over 2}\| 
\leq &  2F \DZF^{1\over 2} \lambda^{1-a}  \|\phi(\TK) \GL(\TK)\|.
\end{align*}
Using the spectral theorem, with \eqref{eq:TKBound}, we get
\begin{align*}
\| \TXL^{{1\over 2} -a }\RL(\TX) \phi(\TK)  \GL(\TK)  \TK^{1\over 2}\| 
\leq &  2F \DZF^{1\over 2} \lambda^{1-a}  \sup_{u \in [0,\kappa^2]} |\GL(u) \phi(u)|.
\end{align*}
When $0<u \leq \lambda,$ as $\phi(u)$ is non-decreasing, $\phi(u) \leq \phi(\lambda)$. Applying \eqref{eq:GLproper1},  we have 
$$\GL(u) \phi(u) \leq E \phi(\lambda)\lambda^{-1}. 
$$
When $\lambda<u\leq \kappa^2,$
following from Lemma \ref{lem:operMon}, we have that there is a $c_{\phi}'\geq 1$, which depends only on $\phi$, $\kappa^2$ and $b$, such that 
$$
{\phi(u)  u^{-1}} \leq c_{\phi}' \phi(\lambda)\lambda^{-1}.
$$
Then, combing with \eqref{eq:GLproper1}, 
$$
\GL(u) \phi(u) = \GL(u) u \phi(u) u^{-1}
 \leq E c_{\phi}' \phi(\lambda)\lambda^{-1} .
$$
Therefore, for all $0<u \leq \kappa^2,$
$\GL(u)\phi(u) \leq E c_{\phi}' \phi(\lambda) \lambda^{-1}$ and consequently, 
\begin{align*}
\| \TXL^{{1\over 2} -a }\RL(\TX) \phi(\TK)  \GL(\TK)  \TK^{1\over 2}\| 
\leq & 2Ec_{\phi}' F \DZF^{1\over 2} \lambda^{-a}\phi(\lambda).
\end{align*}
Introducing the above into \eqref{eq:term2Interm},
we get
\be\label{eq:term2A}
\| \TXL^{{1\over 2} - a}\RL(\TX)  \EPSRA\|_{\HK} \leq 2Ec_{\phi}' FR \DZF^{1\over 2}\lambda^{-a}\phi(\lambda).
\ee
{\it Case 2: $\phi(\cdot)$ is Lipschitz continuous with constant $1$.} \\
By the triangle inequality, we have
\begin{align*}
&\| \TXL^{{1\over 2} -a }\RL(\TX) \phi(\TK)  \GL(\TK)  \TK^{1\over 2}\| \\
\leq& \| \TXL^{{1\over 2} -a }\RL(\TX) \phi(\TX)  \GL(\TK)  \TK^{1\over 2}\| + \| \TXL^{{1\over 2} -a }\RL(\TX) (\phi(\TK) -\phi(\TX)) \GL(\TK)  \TK^{1\over 2}\| \\
\leq & \| \RL(\TX) \phi(\TX)\| \|\TXL^{{1\over 2} -a} \TKL^{a-{1\over 2}}\| \|\TKL^{{1\over 2}-a}  \GL(\TK)  \TK^{1\over 2}\| +  \| \TXL^{{1\over 2} -a }\RL(\TX)\| \|\phi(\TK) -\phi(\TX)\|_{HS}  \|\GL(\TK)  \TK^{1\over 2}\|.
\end{align*}
Since $\phi(u)$ is Lipschitz continuous with constant $1$ and $\phi(0)=0$, then according to Lemma \ref{lem:operLips},
$
\|\phi(\TK) - \phi(\TX)\|_{HS} \leq  \|\TK - \TX \|_{HS}.
$
It thus follows that
\begin{align*}
&\| \TXL^{{1\over 2} -a }\RL(\TX) \phi(\TK)  \GL(\TK)  \TK^{1\over 2}\| \\
\leq & \| \RL(\TX) \phi(\TX) \| \|\TXL^{{1\over 2}-a} \TKL^{a-{1\over 2}}\| \|\TKL^{{1\over 2}-a}  \GL(\TK)  \TK^{1\over 2}\| +  \| \TXL^{{1\over 2} -a }\RL(\TX)\| \|\TK -\TX  \|_{HS} \|\GL(\TK)  \TK^{1\over 2}\|\\
\leq & c_g \phi(\lambda)\|\TXL^{{1\over 2}-a} \TKL^{a-{1\over 2}}\| \|\TKL^{{1\over 2}-a}  \GL(\TK)  \TK^{1\over 2}\| +  \| \TXL^{{1\over 2} -a }\RL(\TX)\| \DZT \|\GL(\TK)  \TK^{1\over 2}\|,
\end{align*}
where for the last inequality, we used \eqref{eq:sorConReslt} to bound $\| \RL(\TX) \phi(\TX) \|$: 
$$\| \RL(\TX) \phi(\TX) \| \leq \sup_{u \in [0,\kappa^2]} | \RL(u) \phi(u) | \leq c_g \phi(\lambda).$$
 Applying Lemma \ref{lem:operProd} which implies 
$$
\|\TXL^{{1\over 2}-a} \TKL^{a-{1\over 2}}\|= \|\TXL^{{1\over 2}(1-2a)} \TKL^{-{1\over 2}(1-2a)}\| \leq \|\TXL^{{1\over 2}} \TKL^{-{1\over 2}}\|^{1-2a}\leq \DZF^{{1\over 2} -a}, 
$$
we get
\begin{align}\label{eq:intermA1}
\| \TXL^{{1\over 2} -a }\RL(\TX) \phi(\TK)  \GL(\TK)  \TK^{1\over 2}\| \leq  c_g \phi(\lambda)\DZF^{{1\over 2} -a} \|\TKL^{{1\over 2}-a}  \GL(\TK)  \TK^{1\over 2}\| +  \| \TXL^{{1\over 2} -a }\RL(\TX)\| \DZT \|\GL(\TK)  \TK^{1\over 2}\|.
\end{align}
By the spectral theorem and \eqref{eq:GLproper1}, with \eqref{eq:TKBound} and $0\leq a\leq {1\over 2}$, we have
\be\label{eq:intermd1}
\|\GL(\TK)  \TK^{1\over 2}\| \leq \sup_{u \in [0,\kappa^2]} |u^{1\over 2} \GL(u)| \leq E \lambda^{-{1\over 2}} \quad \mbox{and}
\ee
$$\|\TKL^{{1\over 2}-a}  \GL(\TK)  \TK^{1\over 2}\| \leq
 \sup_{u \in [0,\kappa^2]} (u^{{1\over 2}-a} + \PRegPar^{{1\over 2}-a}) |\GL(u)| u^{1\over 2} \leq 2E \lambda^{-a} .
$$
Similarly, by \eqref{eq:GLproper4}, with \eqref{eq:TXbound}, 
$$\| \TXL^{{1\over 2} -a }\RL(\TX)\| \leq \sup_{u \in [0,\kappa^2]} (u^{{1\over 2} -a}+ \PRegPar^{{1\over 2} -a}) |\RL(u)| \leq 2F\lambda^{{1\over 2} -a}.
$$
Therefore, following from the above three estimates and 
\eqref{eq:intermA1}, we get
\begin{align}\label{eq:intermA3}
\| \TXL^{{1\over 2} -a }\RL(\TX) \phi(\TK)  \GL(\TK)  \TK^{1\over 2}\| \leq  2c_g \phi(\lambda)\DZF^{{1\over 2} -a} E\lambda^{-a}+  2EF \lambda^{-a} \DZT.
\end{align}
Introducing the above into \eqref{eq:term2Interm},
we get
\begin{align}\label{eq:term2B}
\| \TXL^{{1\over 2} -a }\RL(\TX) \EPSRA\|_{\HK} \leq  2ER\lambda^{-a}(c_g \phi(\lambda)\DZF^{{1\over 2} -a} +  F  \DZT) .
\end{align}
Applying \eqref{eq:term2B} (or \eqref{eq:term2A}) and \eqref{eq:term1} into \eqref{eq:binterm}, by a direct calculation, we get
\begin{align*}
&\|\LK^{-a} (\EESRA  - \FH) \|_{\rho} 
 \leq  \DZF^{{1\over 2} -a}2 E\lambda^{-a} 
(\DZF^{1\over 2} \DZS +  \DZF^{1\over 2} R (c''_{\phi}  + c_g)  \phi(\lambda) + FR\DZT) + c_g R\phi(\lambda) \lambda^{-a}.
\end{align*}
Here, $c''_{\phi} = c_{\phi}' F$ if $\phi$ is operator monotone or $c''_{\phi}=c_g$ if $\phi$ is Lipschitz continuous with constant $1$.  Introducing with $\DZF$, $\DZS$ and $\DZT$, by a direct calculation and $\lambda\leq 1$, one can prove the first part of the theorem
with
$$\tilde{C}_1 = 2 E C_1C_4^{1-a}, \quad \tilde{C}_2 = 2E\sqrt{C}_3 C_4^{1-a} + 12\kappa^2 EFRC_4^{{1\over 2}-a}, \quad \mbox{and}$$
$$\tilde{C}_3 = 2E\sqrt{C}_2 C_4^{1-a} + c_g R + 2ER C_4^{1-a}(c_{\phi}'' + c_g).
$$
{\it Case 3: $\phi = \psi \vartheta$, where $\psi$ is operator monotone and $\vartheta$ is Lipschitz continuous with constant $1$.} \\
Since $\phi = \vartheta \psi,$ we can rewrite $\phi(T)$ as
$$
\phi(\TX) + (\vartheta(\TK) - \vartheta(\TX)) \psi(\TK)  + \vartheta(\TX) (\psi(\TK) - \psi(\TX)).
$$
Thus, together with the triangle inequality,  
\begin{align}
&\| \TXL^{{1\over 2} -a }\RL(\TX) \phi(\TK)  \GL(\TK)  \TK^{1\over 2}\| \nonumber \\
\leq&
\| \TXL^{{1\over 2} -a }\RL(\TX) \phi(\TX)  \GL(\TK)  \TK^{1\over 2}\|  + \| \TXL^{{1\over 2} -a }\RL(\TX) (\vartheta(\TK) - \vartheta(\TX)) \GL(\TK)  \TK^{1\over 2}\| \|\psi(\TK)\| \nonumber\\
&+ 
\| \TXL^{{1\over 2} -a }\RL(\TX)\vartheta(\TX)(\psi(\TK) - \psi(\TX))  \GL(\TK)  \TK^{1\over 2}\| \label{eq:intermd4}. 
\end{align}
Following the same argument as that for \eqref{eq:intermA3}, we know that
\be\label{eq:intermd2}
\| \TXL^{{1\over 2} -a }\RL(\TX) \phi(\TX)  \GL(\TK)  \TK^{1\over 2}\| \leq 2c_gE \phi(\lambda)\DZF^{{1\over 2} -a} \lambda^{-a},
\ee
and 
\be\label{eq:intermd3}
\| \TXL^{{1\over 2} -a }\RL(\TX) (\vartheta(\TK) - \vartheta(\TX)) \GL(\TK)\TK^{1\over 2} \|  \leq 2EF \lambda^{-a}  \DZT.
\ee

As the quality of $\GL$ covers $\vartheta(u) u^{{1\over 2}-a},$  
applying the spectral theorem and Lemma \ref{lem:sorConReslt}, we get
$$
\| \TXL^{{1\over 2} -a }\RL(\TX)\vartheta(\TX)\| \leq \sup_{u \in [0,\kappa^2]} (u+ \PRegPar)^{{1\over 2} -a} \RL(u) \vartheta(u) \leq c_g'\vartheta(\lambda)(\lambda^{{1\over 2}-a} + \PRegPar^{{1\over 2}-a}).
$$
Since $\psi$ is operator monotone on $[0,b]$ where $b>\kappa^2$, we know from Lemma \ref{lem:operMon} that there exists a positive constant $c_{\psi}<\infty$ depending on $b-\kappa^2,$ such that
$$
\|\psi(\TX) - \psi(\TK)\| \leq c_{\psi} \psi(\|\TK - \TX\|).
$$
If $\sqrt{n} \geq 6 \log{6 \over \delta},$
as $\psi$ is non-decreasing, following from \eqref{eq:statEstA}, we have $\psi(\|\TK - \TX\|) \leq \psi(\|\TK - \TX\|_{HS}) \leq \psi(\DZT)$ and thus 
$$
\|\psi(\TX) - \psi(\TK)\| \leq c_{\psi} \psi(\DZT) \leq c_{\psi}c_{\psi}' \psi(n^{-1/2}) {6\kappa^2 \log{6 \over \delta}},
$$
where for the last inequality, we used Lemma \ref{lem:operMon}. If $\sqrt{n} \leq 6 \log{6 \over \delta},$ then as $\|\TK - \TX\| \leq \max(\|\TK\|, \|\TX\|) \leq \kappa^2,$
$$
\|\psi(\TX) - \psi(\TK)\| \leq  c_{\psi}\psi(\kappa^2) \leq {c_{\psi}\psi(\kappa^2) 6\log {6\over \delta} } {1\over \sqrt{n}}  \leq c_{\psi}'{c_{\psi}6\kappa^2 \log {6\over \delta} } \psi(n^{-1/2}),
$$
where for the last inequality, we used Lemma \ref{lem:operMon}. 
Therefore, following from the above analysis and \eqref{eq:intermd1},
\begin{align*}
&\| \TXL^{{1\over 2} -a }\RL(\TX)\vartheta(\TX)(\psi(\TK) - \psi(\TX))  \GL(\TK)  \TK^{1\over 2}\| \\\leq&  \| \TXL^{{1\over 2} -a }\RL(\TX)\vartheta(\TX)\| \|\psi(\TK) - \psi(\TX)\| \| \GL(\TK)  \TK^{1\over 2}\| \\
\leq& 12c_g' c_{\psi}c_{\psi}' E  {\kappa^2 \log{6 \over \delta}} \lambda^{-a} \vartheta(\lambda)  \psi(n^{-1/2}) .
\end{align*}
Introducing the above estimate, \eqref{eq:intermd2} and \eqref{eq:intermd3} into \eqref{eq:intermd4}, with $\|\psi(\TK)\| \leq \psi(\kappa^2)$ (since $\psi$ is operator monotone and \eqref{eq:TKBound}),
we conclude that
\begin{align*}
&\| \TXL^{{1\over 2} -a }\RL(\TX) \phi(\TK)  \GL(\TK)  \TK^{1\over 2}\| \\
\leq&  2\lambda^{-a} \left(c_gE \phi(\lambda)\DZF^{{1\over 2} -a} + EF \DZT\psi(\kappa^2)+ 6\kappa^2 c_g' c_{\psi} c_{\psi}' E \vartheta(\lambda) \psi(n^{-{1\over 2}}) \log {6\over \delta}\right).
\end{align*}
Introducing the above into \eqref{eq:term2Interm}, we get
\begin{align*}
\| \TXL^{{1\over 2} -a }\RL(\TX) \EPSRA\| 
\leq 2 \lambda^{-a} \left(c_gE \phi(\lambda)\DZF^{{1\over 2} -a} + EF\psi(\kappa^2) \DZT+ 6\kappa^2 c_g' c_{\psi} c_{\psi}' E \vartheta(\lambda) \psi(n^{-{1\over 2}}) \log {6\over \delta}\right)R. 
\end{align*}
Combining the above and \eqref{eq:term1} with \eqref{eq:binterm}, by a direct calculation,  we get
\begin{align*}
&\|\LK^{-a} (\EESRA  - \FH) \|_{\rho} \\
\leq&  \lambda^{-a} \Big( 2E\DZF^{1-a}(\DZS + 2c_gR \phi(\lambda)) + c_g R\phi(\lambda) + 2EFR \psi(\kappa^2) \DZF^{{1\over 2}-a} \DZT 
\Big.\\
&\Big.+ 12\kappa^2 c_g'ER c_{\psi} c_{\psi}' \DZF^{{1\over 2}-a}\vartheta(\lambda)\psi(n^{-{1\over 2}}) \log {6\over \delta} \Big).
\end{align*}
Introducing with $\DZF$, $\DZS$ and $\DZT$, by a simple calculation, with $\lambda\leq 1$, we can prove the second part of the theorem with
$$\tilde{C}_4 = 2 E C_4^{1-a} \sqrt{C_3} + 12 EFR\psi(\kappa^2) \kappa^2 C_4^{{1\over 2}-a}, \quad \tilde{C}_5 = 2 E C_4^{1-a}(3c_g R + \sqrt{C_2}), 
$$
$$\mbox{and}\quad \tilde{C}_6 = 12\kappa^2 c_g' c_{\psi} c_{\psi}' ER C_4^{{1\over 2}-a}.$$
\end{proof}

\begin{proof}[Proof of Corollary \ref{cor:generalSoucr}]
	Let  $\theta$ be such that $\PRegPar = n^{\theta-1}$. As
	$\Theta(u)$ is non-decreasing, $\Theta(0)=0$, $\Theta(1) = 1$ and that $\lambda$ satisfies $\Theta(\lambda) = n^{-1}$, then $0\leq \lambda \leq 1.$ 
	 Moreover,  
	as that $\phi(\lambda)\lambda^{-\zeta}$ is non-decreasing which implies
	\be\label{eq:B}
	   \left( {\phi(1) \over \sqrt{n} \phi(\lambda) \lambda^{-\zeta}}\right)^2 \geq {1 \over n},
	\ee
	and that $$ \lambda^{\gamma+2\zeta} = \left( {\phi(1) \over \sqrt{n} \phi(\lambda) \lambda^{-\zeta}}\right)^2,
	$$
	then $\lambda \geq n^{-{1\over 2\zeta+\gamma}}$. Thus, with $2\zeta+\gamma >1$,  
	$\theta = \log_{n} \lambda + 1 \geq -{1\over 2\zeta+\gamma} + 1>0.
	$
	Also, $\theta \leq 1$ as $\lambda \leq 1.$ Applying Part 1) of Theorem \ref{thm}, and noting that by $2\zeta+\gamma >1$, $1 \geq \lambda\geq n^{-{1\over 2\zeta+\gamma}} $ and \eqref{eq:B},
	$$ {1\over n \lambda^{{1\over 2 } }} \leq {1\over \sqrt{n}} \leq  {1\over \sqrt{n\lambda^{\gamma}}} = {\phi(\lambda)\over \phi(1)}, \quad {1\over n\lambda^{1-\zeta}} \leq {1\over \sqrt{n\lambda^{\gamma}}} \  (\mbox{if } 2\zeta \leq 1).
	$$
	 we can prove the first desired result. The second desired result can be proved by using Part 2) of Theorem \ref{thm}, the above estimates, as well as $\psi(n^{-1/2}) \leq \psi(\lambda)$ (since $\psi$ is non-decreasing).
\end{proof}

\begin{proof}[Proof of Corollary \ref{cor}]
	If $\zeta\leq 1,$ then
	$\phi $ is operator monotone \cite[Theorem 1 and Example 1]{mathe2002moduli}. If $\zeta \geq 1,$ then $\phi$ is Lipschitz continuous with constant $1$ over  $[0,\kappa^2]$. Applying Part 1) of Theorem \ref{thm}, one can prove the desired results.
	\end{proof}

\begin{proof}[Proof of Corollary \ref{cor:3}]
The proof can be done by using Corollary \ref{cor} with simple arguments. For notional simplicity,  we let 
	\bea
\Lambda_n =  \begin{cases} n^{-{(\zeta-a) \over 2\zeta+\gamma}}  & \mbox{ if } 2\zeta+\gamma>1, \\
	n^{-(\zeta-a)}  \left( 1 \vee \log n^{\gamma}\right)^{(1-a)}
	& \mbox{ if } 2\zeta+\gamma\leq 1.
\end{cases} 
\eea
\\
1) 	Using the fact that for any non-negative random variable $\xi$, $\mE[\xi] = \int_{t\geq 0} \Pr(\xi \geq t)d t,$ and Corollary \ref{cor}, for any $q\in \mN_+$
$$
\mE\|\LK^{-a} (\EESRA  - \FH) \|_{\rho}^q \leq C \int_{t\geq 0} \exp \left\{ \left(-{t \over C \Lambda_n^q}\right)^{1 \over q(2-a)} \right\} d t \leq C \Lambda_n^q.
$$
2) By Corollary \ref{cor}, we have, with $\delta_n = n^{-2}$,
$$
\sum_{n=1}^{\infty} \Pr\left( n^{ \zeta-a - \epsilon \over 1 \vee (2\zeta+\gamma)}\|\LK^{-a} (\EESRA  - \FH) \|_{\rho}  > C n^{ \zeta-a - \epsilon \over 1 \vee (2\zeta+\gamma)}\Lambda_n \log^{2-a} {6\over \delta_n}\right) \leq \sum_{n=1}^{\infty} \delta_n^2 < \infty. 
$$ 
Note that $C n^{ \zeta-a - \epsilon \over 1 \vee (2\zeta+\gamma)}\Lambda_n \log^{2-a} {6\over \delta_n} \to 0$ as $n\to \infty.$ Thus,
applying the Borel-Cantelli lemma, one  can prove Part 2). \\
3)  Following the argument from the proof of Corollary \ref{cor}, one can prove Part 3).
	\end{proof}

\section*{Acknowledgment}
This manuscript version is made available under the CC-BY-NC-ND 4.0 license.
JL and VC's work was supported in
part by Office of Naval Research (ONR) under grant agreement number N62909-17-1-2111, in part by Hasler Foundation Switzerland under grant agreement number 16066, and 
 in part by the European Research Council (ERC) under the European Union's Horizon 2020 research and innovation program (grant agreement number 725594).

\bibliography{distributed_alg_sgm_sra}
\bibliographystyle{abbrv}
\appendix
{\section{List of Notations} \label{sec:notations}}
{\footnotesize
	\begin{center}
		\begin{longtable}{ c | l  }
			\hline			
			Notation &  Meaning  \\ \hline
			$\HK$ & the input space - separable Hilbert space\\
			
			
			$\rho$, $\rho_X$ & the fixed  probability measure on $\HK \times \mR$, the induced marginal measure of $\rho$ on $\HK$
			\\
			
			$\rho(\cdot | x)$ &  the conditional probability measure on $\mR$ w.r.t. $x\in \HK$ and $\rho$ \\
			
			$\HR$ & the hypothesis space, $\{f: \HK \to  \mR| \exists \omega \in \HK \mbox{ with } f(x) = \la \omega, x \ra_{\HK}, \rho_X \mbox{-almost surely}\}.$  \\
			
			$n$ & the sample size\\
			
			$\Samples$ & the whole samples $\{{z}_i\}_{i=1}^n$, where each $z_i$ is i.i.d. according to $\rho$\\
			
			$\by$ & the vector of sample outputs, $(y_1,\cdots,y_n)^{\top}$ \\
			
			$\bx$ & the set of sample outputs, $\{x_1,\cdots,x_n\}$ \\

			$\mcE$ & the expected risk defined by \eqref{expectedRisk}\\
			
				$\LR$ & the Hilbert space of square integral functions from $\HK$ to $\mR$ with respect to $\rho_X$ \\ 
			
			$f_{\rho}$ & the regression function defined  \eref{regressionfunc} \\
			
			$\kappa^2$ & the constant from the bounded assumption \eqref{boundedKernel} on the input space $\HK$ \\

			$\IK$ & the linear map from $\HK \to \LR$ defined by $\IK \omega = \la \omega, \cdot \ra_{\HK}$  \\
			
			$\IK^*$ & the adjoint operator of $\IK$: $\IK^* f = \int_{X} f(x) x d\rho_{X}(x) $ \\
			
			$\LK$ & the operator from $\LR $ to $\LR$, $\LK(f) = \IK \IK^*f =\int_{X}  \la x, \cdot\ra_{\HK} f(x) \rho_{X}(x)$\\
			
			$\TK$ & the covariance operator from $\HK$ to $\HK$, $\TK = \IK^* \IK = \int_{X} \la \cdot, x \ra x d\rho_{X}(x)$ \\

			$\SX$ & the sampling operator from $\HK$ to $\mR^{n}$, $(\SX \omega)_i = \la \omega, x_i \ra_{\HK}, i \in \{1,\cdots,n\}$ \\
			
			$\SX^*$ & the adjoint operator of  $\SX$, $\SX^* \mathbf{y} = {1 \over n} \sum_{i=1}^n y_i x_i$\\
			
			$\TX$ & the empirical covariance operator, $\TX = \SX^* \SX = {1 \over n} \sum_{i=1}^n \la \cdot, x_i\ra x_i$\\
			
			$\FH$ & the projection of $\FR$ onto the closure of $\HR$ in $\LR$ \\

			$\GL(\cdot)$ &  the filter function of the regularized algorithm from Definition \ref{def} \\
			
			$\tau$ & the qualification of the filter function $\GL$\\
			
			$E,F_{\tau}$ & the constants related to the filter function $\GL$ from \eqref{eq:GLproper1} and \eqref{eq:GLproper4} \\

			$\lambda$ & a regularization parameter $\lambda>0$ \\

			$\ESRA$ & an estimated vector defined by \eqref{eq:ESRA} \\
			
			$\EESRA$  & an estimated function defined by \eqref{eq: EESRA} \\
			
			$M, Q$ & the positive constants from 
			Assumption \eqref{noiseExp} \\
			
			$B$ & the constant from \eqref{eq:FHFR} \\
			
			$\phi,R$& the function and the parameter related to the `regularity' of $\FH$ (see Assumption \ref{as:regularity}) \\
			
			$\gamma, c_{\gamma}$ & the parameters related to the effective dimension (see Assumption \ref{as:eigenvalues}) \\
			
			$\{\sigma_i\}_i$ & the sequence of eigenvalues of $\LK$ \\
			
			$\psi, \vartheta$ & the functions from Part 2 of Theorem \ref{thm}, $\phi = \psi \vartheta$ \\

			
			$\TKL$, &  $\TKL = \TK + \PRegPar$\\
			
			$\TXL$, &  $\TXL = \TX  +\PRegPar$\\

			
			
			


$\zeta$ & the parameter related to the Holder source condition on $\FH$  (see \eqref{eq:HSCB})\\
			
$\RL(u)$ & $ = 1 - \GL(u)u$ \\

$\mcN(\lambda)$ & $= \tr (\TK(\TK + \lambda)^{-1})$\\

$c_g$ & the constant from Lemma \ref{lem:sorConReslt} \\

$\omega_{\lambda}$ & the population vector defined by \eqref{eq:popFunc} \\ 
			
			$a_{n,\delta,\gamma}(\theta)$ & the quantity defined by \eqref{eq:aa}\\


			\hline  
		\end{longtable}
	\end{center}
\end{document}